\documentclass[10pt,twocolumn,letterpaper]{article}

\usepackage{cvpr}              

\usepackage{graphicx}
\usepackage{amsmath}
\usepackage{amssymb}
\usepackage{booktabs}
\usepackage{graphicx}
\usepackage{url}
\usepackage{bbding}
\usepackage{pifont}
\usepackage{pbox}
\usepackage{booktabs}
\usepackage[dvipsnames]{xcolor}
\usepackage{float}
\usepackage{multirow}

\usepackage{wrapfig,lipsum,booktabs}
\usepackage{color}

\usepackage{comment}
\usepackage{amsmath} 

\definecolor{applegreen}{rgb}{0.0, 0.5, 0.0}
\newcommand{\name}{\texttt{Consistent-Teacher}~}

\usepackage[pagebackref,breaklinks,colorlinks]{hyperref}

\usepackage[capitalize]{cleveref}
\crefname{section}{Sec.}{Secs.}
\Crefname{section}{Section}{Sections}
\Crefname{table}{Table}{Tables}
\crefname{table}{Tab.}{Tabs.}

\def\cvprPaperID{7649} 
\def\confName{CVPR}
\def\confYear{2023}

\begin{document}

\title{Consistent-Teacher: Towards Reducing Inconsistent Pseudo-targets in Semi-supervised Object Detection}


\author{Xinjiang Wang$^1$$^*$\quad Xingyi Yang$^{3}$$^{* \dag}$\quad Shilong Zhang$^2$, Yijiang Li$^1$$^\ddag$\\ Litong Feng$^1$\quad  Shijie Fang$^4$$^\ddag$\quad Chengqi Lyu$^2$\quad Kai Chen$^{2}$\quad Wayne Zhang$^1$ \\\newline
{\small $^1$SenseTime Research\quad 
$^2$Shanghai AI Laboratory\quad 
$^3$National University of Singapore\quad 
$^4$Peking University} \\
{\small\texttt{wangxinjiang@sensetime.com, xyang@u.nus.edu}} }
\maketitle
\def\thefootnote{*}\footnotetext{Equally contributed.}\def\thefootnote{\arabic{footnote}}
\def\thefootnote{\ddag}\footnotetext{Work done during internship at SenseTime.}\def\thefootnote{\arabic{footnote}}
\def\thefootnote{\dag}\footnotetext{Work done during internship at Shanghai AI Laboratory.}\def\thefootnote{\arabic{footnote}}
\begin{abstract}
In this study, we dive deep into the inconsistency of pseudo targets in semi-supervised object detection (SSOD). Our core observation is that the oscillating pseudo-targets undermine the training of an accurate detector. It injects noise into the student's training, leading to severe overfitting problems. Therefore, we propose a systematic solution, termed \name, to reduce the inconsistency.
First, adaptive anchor assignment~(ASA) substitutes the static IoU-based strategy, which enables the student network to be resistant to noisy pseudo-bounding boxes. Then we calibrate the subtask predictions by designing a 3D feature alignment module~(FAM-3D). It allows each classification feature to adaptively query the optimal feature vector for the regression task at arbitrary scales and locations. Lastly, a Gaussian Mixture Model (GMM) dynamically revises the score threshold of pseudo-bboxes, which stabilizes the number of ground truths at an early stage and remedies the unreliable supervision signal during training. \name provides strong results on a large range of SSOD evaluations. It achieves 40.0 mAP with ResNet-50 backbone given only 10\% of annotated MS-COCO data, which surpasses previous baselines using pseudo labels by around 3 mAP. When trained on fully annotated MS-COCO with additional unlabeled data, the performance further increases to 47.7 mAP. 
 Our code is available at \url{https://github.com/Adamdad/ConsistentTeacher}.
\end{abstract}

\section{Introduction}
\label{sec:intro}

The goal of semi-supervised object detection (SSOD)~\cite{jeong2021interpolation,chen2021temporal,sohn2020simple,jeong2019consistency,liu2021unbiased,jeong2019consistency,xu2021end,zhou2021instant,chen2022label,liu2022unbiased,li2022pseco,zhou2022dense} is to facilitate the training of object detectors  with the help of a large amount of unlabeled data. The common practice is first to train a teacher model on the labeled data and then generate pseudo labels and boxes on unlabeled sets, which act as the ground truth (GT) for the student model. Student detectors, on the other hand, are anticipated to make consistent predictions regardless of network stochasticity~\cite{xie2020self} or data augmentation~\cite{jeong2019consistency,sohn2020simple}. In addition, to improve pseudo-label quality, the teacher model is updated as a moving average~\cite{liu2021unbiased,xu2021end,zhou2021instant} of the student parameters. 

In this study, we point out that the performance of semi-supervised detectors is still largely hindered by the inconsistency in pseudo-targets. \textbf{Inconsistency} means that the pseudo boxes may be highly inaccurate and vary greatly at different stages of training. 
As a consequence, inconsistent oscillating bounding boxes~(bbox) bias SSOD predictions with accumulated error. Different from semi-supervised classification, SSOD has one extra step of assigning a set of pseudo-bboxes to each RoI/anchor as dense supervision. Common two-stage~\cite{sohn2020simple, liu2021unbiased, xu2021end} and single-stage ~\cite{zhang2022s4od, chen2022dense} SSOD networks adopt static criteria for  anchor assignment, \emph{e.g.} IoU score or centerness. It is observed that the static assignment is sensitive to noise in the bounding boxes predicted by the teacher, as a small perturbation in the pseudo-bboxes might greatly affect the assignment results. It thus leads to severe overfitting on unlabeled images.

\begin{figure*}
    \centering
    \includegraphics[width=\linewidth]{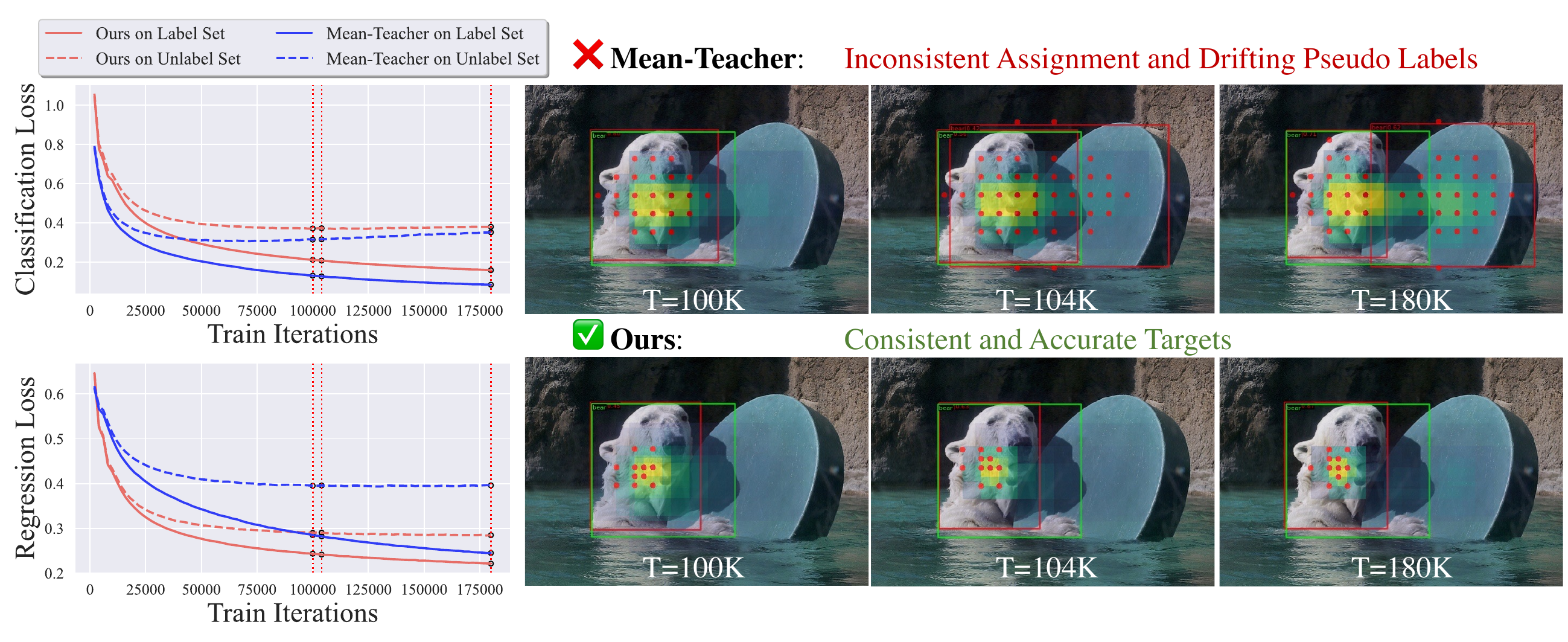}
    \vspace{-5mm}
    \caption{Illustration of inconsistency problem in SSOD on COCO 10~\% evaluation. (Left) We compare the training losses between the Mean-Teacher and our \name. In Mean-Teacher, inconsistent pseudo targets lead to overfitting on the classification branch, while regression losses become difficult to converge. In contrast, our approach sets consistent optimization objectives for the students, effectively balancing the two tasks and preventing overfitting. (Right) Snapshots for the dynamics of pseudo labels and assignment. The \textcolor{applegreen}{Green} and \textcolor{red}{Red} bboxes refer to the ground-truth and pseudo bbox, respectively, for the polar bear. \textcolor{red}{Red dots} are the assigned anchor boxes for the pseudo label. The heatmap indicates the dense confidence score predicted by the teacher~(brighter the larger). A nearby board is finally misclassified as a polar bear in the baseline while our adaptive assignment prevents overfitting.}
    \label{fig:motivation}
     \vspace{-3mm}
\end{figure*}
To verify this phenomenon, we train a single-stage detector with standard IoU-based assignment on MS-COCO 10\% data. 
As shown in Fig.~ (\ref{fig:motivation}), a small change in the teacher's output results in strong noise in the boundaries of pseudo-bboxes, causing erroneous targets to be associated with nearby objects under static IoU-based assignment.
This is because some inactivated anchors are falsely assigned positive
in the student network.
Consequently, the network overfits as it produces inconsistent labels for neighboring objects. The overfitting is also observed in the classification loss curve on unlabeled images\footnote{All GT bboxes on unlabeled data are only used to calculate the loss value but not for updating the parameters.}. 


Through dedicated investigation, We find that one important factor that leads to the drifting pseudo-label is the mismatch between classification and regression tasks. 
Typically, only the classification score is used to filter pseudo-bboxes in SSOD. However, confidence does not always indicate the quality of the bbox~\cite{xu2021end}. 
Two anchors with similar scores, as a result, can have significantly different predicted pseudo-bboxes, leading to more false predictions and label drifting. Such phenomenon is illustrated in Fig.~(\ref{fig:motivation}) with the varying pseudo-bboxes of the MeanTeacher around $T=104K$. Therefore, the mismatch between the quality of a bbox and its confidence score would result in noisy pseudo-bboxes, which in turn exacerbates the label drifting.

The widely-employed hard threshold scheme also causes threshold inconsistencies in pseudo labels. Traditional SSOD methods~\cite{sohn2020simple,liu2021unbiased,xu2021end} utilize a static threshold on confidence score for student training. However, the threshold serves as a hyper-parameter, which not only needs to be carefully tuned but should also be dynamically adjusted in accordance with the model's capability at different time steps. In the Mean-Teacher\cite{tarvainen2017mean} paradigm, the number of pseudo-bboxes may increase from too few to too many under a hard threshold scheme, which incurs inefficient and biased supervision for the student.

Therefore, we propose \name in this study to address the inconsistency issues. First, we find that a simple replacement of the static IoU-based anchor assignment by cost-aware adaptive sample assignment~(ASA)~\cite{ge2021ota, ge2021yolox} greatly alleviates the effect of inconsistency in dense pseudo-targets. During each training step, we calculate the matching cost between each pseudo-bbox with the student network's predictions. Only feature points with the lowest costs are assigned as positive. It reduces the mismatch between the teacher's high-response features and the student's assigned positive pseudo targets, which inhibits overfitting.

Then, we calibrate the classification and regression tasks so that the teacher's classification confidence provides a better proxy of the bbox quality. It produces consistent pseudo-bboxes for anchors of similar confidence scores, and thus the oscillation in pseudo-bbox boundaries is reduced. 
Inspired by TOOD~\cite{feng2021tood}, we  propose a 3-D feature alignment module~(FAM-3D) that allows classification features to sense and adopt the best feature in its neighborhood for regression. Different from the single scale searching, FAM-3D reorders the features pyramid for regression across scales as well.
In this way, a unified confidence score accurately measures the quality of classification and regression with the improved alignment module and ultimately brings consistent pseudo-targets for the student in SSOD.

As for the threshold inconsistency in pseudo-bboxes, we apply Gaussian Mixture Model (GMM) to generate an adaptive threshold for each category during training. 
We consider the confidence scores of each class as the weighted sum of positive and negative distributions and predict the parameters of each Gaussian with maximum likelihood estimation.
It is expected that the model will be able to adaptively infer the optimal threshold at different training steps so as to stabilize the number of positive samples. 

The proposed \name greatly surpasses current SSOD methods. Our approach reaches 40.0 mAP with 10\% of labeled data on MS-COCO, which is \~3 mAP ahead of the state-of-the-art~\cite{zhou2022dense}. When using the 100\% labels together with extra unlabeled MS-COCO data, the performance is further boosted to 47.7 mAP. The effectiveness of \name is also testified on other ratios of labeled data and on other datasets as well. 
Concretely, the paper contributes in the following aspects. 
\begin{itemize}
  \setlength\itemsep{0.0em}
    \item We provide the first in-depth investigation of the inconsistent target problem in SSOD, which incurs severe overfitting issues. 
    \item We introduce an adaptive sample assignment to stabilize the matching between noisy pseudo-bboxes and anchors, leading to robust training for the student.
    \item We develop a 3-D feature alignment module (FAM-3D) to calibrate the classification confidence and regression quality, which improves the quality of pseudo-bboxes. 
    \item We adopt GMM to flexibly determine the threshold for each class during training. The adaptive threshold evolves through time and reduces the threshold inconsistencies for SSOD. 
    \item \name achieves compelling improvement on a wide range of  evaluations and serves as a new solid baseline for SSOD.
\end{itemize}

\section{Related Work}


\noindent\textbf{Semi-supervised object detection (SSOD).} 
It is a common practice for SSOD to generate pseudo bounding boxes using a teacher model and expect the student detectors to make consistent predictions on augmented input samples\cite{jeong2019consistency,sohn2020simple,yang2022factorizing, liu2021unbiased,xu2021end,zhou2021instant,li2020improving,wang2021data,tang2021humble}. 
Two-stage detectors~\cite{jeong2019consistency, liu2021unbiased,xu2021end} have been dominant in traditional SSOD methods while single-stage detectors have also shown the advantages for their simplicity and higher performance~\cite{zhang2022s4od, chen2022dense,zhou2022dense}.  
In this study, we adopt a single-stage SSOD framework~\cite{zhou2022dense,chen2022dense} and focus on the inconsistency problem.
To resolve the inconsistency issues, we design the adaptive anchor assignment, feature alignment, and GMM-based threshold to improve the label quality.

\noindent\textbf{Label assignment in object detection.} Defining positive and negative sample~\cite{zhang2020bridging} plays a substantial role in object detection. Typical Anchor-based or anchor-free detectors either adopt hard IoU thresholding ~\cite{ren2015faster,cai2018cascade,lin2017feature,lin2017focal,liu2016ssd,dai2016r,redmon2017yolo9000,yang2020scale} or the centerness prior~\cite{redmon2016you, tian2019fcos,kong2020foveabox} as the assigning criterion. 
In contrast, modern detectors have been shifting to adaptive assignment strategies.~\cite{zhang2019freeanchor, ke2020multiple, zhu2020autoassign, kim2020probabilistic, ge2021ota}  
For example, 
PAA~\cite{kim2020probabilistic} adaptively differentiates  the positive anchors and negative ones by fitting the anchor scores distribution. OTA~\cite{ge2021ota} treats the label assignment as an optimal transport problem so that the assignment cost is minimized. 

Although the existing assignment methods are effective, they are limited to fully-supervised settings. In our work, we observe that using static assignment in SSOD induces server inconsistency issues and accumulates errors. We show that a simple cost-ware assignment stabilizes the label noise and significantly improves the performance of SSOD.

\section{\name}
\begin{figure*}[t]
    \centering
    \includegraphics[width=\linewidth]{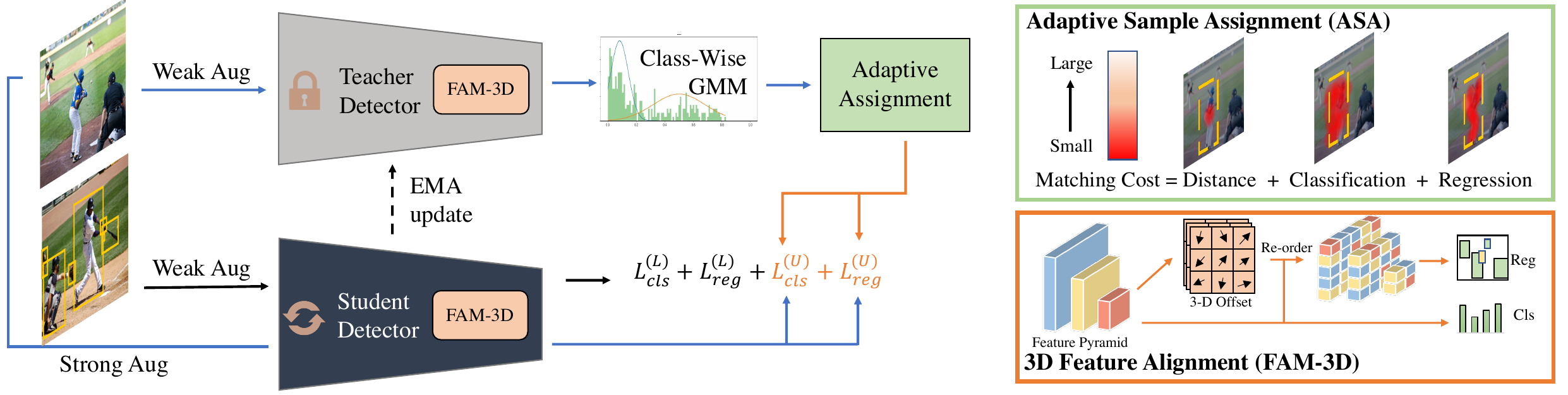}
    \vspace{-5mm}
    \caption{The pipeline of~\name. We design three modules to address the inconsistency in SSOD, where GMM dynamically determines the threshold; 3D feature alignment calibrates regression quality; Adaptive assignment assigns anchor based on matching cost.}
    \label{fig:pipeline}
    \vspace{-4mm}
\end{figure*}

In this section, we elaborate on how our \name works to address the SSOD inconsistencies. It is composed of three key modules, namely Adaptive Sample Assignment, 3D Feature Alignment Module, and Gaussian Mixture-based thresholding. The full pipeline is in Figure~\ref{fig:pipeline}.
\subsection{Baseline Semi-Supervised Detector}

We adopt a general SSOD paradigm as our baseline, namely a Mean-Teacher~\cite{liu2021unbiased,xu2021end,tarvainen2017mean} pipeline with a RetinaNet~\cite{lin2017focal} detector. The teacher model is an exponential moving average~\cite{tarvainen2017mean} of a student detector. Unlabeled images first go through weak augmentations and are fed into the teacher detector to generate pseudo-bboxes. Pseudo-bboxes are then used as supervision for the student network, whose unlabeled images are strongly jittered. 
In the meantime, the student detector takes the labeled images as input to learn discriminative representation for both classification and regression.
Given a labeled set $\mathcal{D}_L = \{\mathbf{x}^{l}_i, \mathbf{y}^{l}_i\}^{N}$ with $N$ samples and an unlabeled set $\mathcal{D}_U = \{\mathbf{x}^u_j\}^{M}$ with $M$ samples, we maintain a teacher detector $f_t(\cdot;\Theta_t)$ and a student detector $f_s(\cdot;\Theta_s)$ that minimize the loss
\begin{small}
\begin{align}
    \begin{split}
        \mathcal{L} = & \frac{1}{N}\sum_i \Big[\mathcal{L}_{cls}\big(f_s(T(\mathbf x_i^l)), \mathbf y_i^l\big) + \mathcal{L}_{reg}\big(f_s(T(\mathbf x_i^l)), \mathbf y_i^l\big)\Big] \\&  +\lambda_u\frac{1}{M}\sum_j \Big[\mathcal{L}_{cls}\big(f_s( T'(\mathbf x_j^u)), \hat{\mathbf {y}}_j^u\big) + \mathcal{L}_{reg}\big(f_s(T'(\mathbf x_j^u)), \hat{\mathbf {y}}_j^u\big)\Big],
    \end{split}
\end{align}
\end{small}
where $T$ and $T'$ stands for weak and strong image transformations, $\mathbf y=\{y_l = (c_l, \mathrm{bbox}_l)\}_{l=1}^L$ is the ground truth (GT) including $L$ bboxes with classification label $c_l$.  $\hat{\mathbf{y}}=f_t(T(\mathbf{x});\Theta_t)$ is the pseudo-bboxes generated by the teacher model.  Teacher parameter is updated as $\Theta_t \leftarrow (1-\gamma) \Theta_t + \gamma \Theta_s$. $\lambda_u$ is a weighting parameter. To ensure a fair comparison, Focal Loss~\cite{lin2017focal} and GIoU loss~\cite{rezatofighi2019generalized} are set for $\mathcal{L}_{cls}$ and $\mathcal{L}_{reg}$ for all models in this study.

\subsection{Consistent Adaptive Sample Assignment}
Each anchor in RetinaNet is assigned as positive only if its IoU with ground truth (GT) bbox is larger than a threshold. 
{Such static label assignment breaks one important property in semi-supervised learning. Take classification as an example, the instance-level pseudo-label satisfies
\begin{equation}
\hat{c} = \mathop{\mathrm{argmin}}_{c}\mathcal L(f_t(\mathbf x^u), c),
\end{equation}
meaning that the pseudo-label $\hat{c}$ should align with its own prediction. }
However, this rule is broken when adopting static anchor assignment to SSOD. That is, the assigned labels for anchors sometimes contradict their own predictions, which is the root of the pseudo-label drifting phenomenon in Fig.~\ref{fig:motivation}.
Therefore, we propose to assign pseudo-bboxes to anchors that minimize their loss
\begin{align}
\label{eq:asa}
\min_{a_1, \cdots, a_N} \sum_n^N \Big[\mathcal{L}_{cls}\big(f_s( \mathbf x^u)_n, \hat{\mathbf {y}}_{a_n}^u)\big) + \mathcal{L}_{reg}\big(f_s(\mathbf x^u)_n, \hat{\mathbf {y}}_{a_n}^u\big)\Big]
\end{align}
where $n$ is the anchor index, and $a_n \in \{1, 2, \cdots, L+1\}$ stands for the assigned pseudo-bbox index from the $L$ predicted bboxes, and the index $L+1$ represents the background label. 
 
A simple solution to Eq.~\ref{eq:asa} is to assign anchors of lowest losses as positive for a pseudo-bbox.
In practice, a matching cost between each anchor\footnote{Our anchor definition generalizes to anchor points in anchor-free and anchor boxes in anchor-based detectors.} and pseudo-bbox is calculated, and the anchors with the lowest costs are considered positive. Given an anchor $n$, the cost between each pseudo-bbox $y_l$ and the prediction $p_n$ from the anchor is calculated as 
\begin{equation}
    \label{eq:cost}
    C_{nl} =\mathcal L_{cls}(p_n, y_l)+ \lambda_{reg} \mathcal L_{reg}(p_n, y_l) + \lambda_{dist}C_{dist},
\end{equation}
where  $\lambda_{reg}$ and $\lambda_{dist}$ are weighting parameters. $C_{dist}$ calculates the distance between the center of anchor $n$ and pseudo-bbox $y_l$, serving as a center prior with a small weighting value ($\lambda_{dist}\sim 0.001$) to stabilize the training. 
With the matching cost for each pseudo-bbox, anchors with top $K$ lowest costs are assigned as positive. 
Since the assignment is made in accordance with the model's detection quality, noise in pseudo-bboxes would then have a negligible impact on the feature points assignment.


We are aware that a similar anchor assignment is adopted in supervised object detection~\cite{ge2021ota,ge2021yolox,carion2020end}, and thus we adopt a unified  assignment for both labeled and unlabeled images.  Despite their similar form, our ASA module addresses the unique pseudo-label shifting issue instead of catering for object variations in supervised settings\cite{ge2021ota}.

\subsection{BBox Consistency via 3-D Feature Alignment}
\label{subsec:subtask_consistency}

In common SSOD frameworks, pseudo-bboxes are generated purely according to classification scores. A high-confidence prediction,  however, does not always guarantee accurate bbox localization~\cite{xu2021end}. It again contributes to the noise in the pseudo-bbox. 
Therefore, inspired by TOOD~\cite{feng2021tood}, we introduce a 3-D Feature Alignment Module (FAM-3D) to calibrate the bbox localization with classification confidence. It allows each classification feature to adaptively locate the optimal feature for the regression task. 

Assuming the feature pyramid is $\mathbf{P}$ with $P(i, j, l)$ indicating the spatial location $(i, j)$ at the $l^{\mathrm{th}}$ pyramid level, we would like to construct a re-sampling function $\mathbf{P}' \leftarrow s(\mathbf{P})$ to rearrange the feature map to conduct the regression task, so that $\mathbf{P}'$ better aligns with the classification features. Different from the single-scale feature re-sampling in~\cite{feng2021tood}, we extend the process to multi-scale feature space, considering the fact that the optimal features for classification and regression could be at different scales~\cite{liu2018path}.

Our feature alignment is realized via a sub-branch in the detection head that predicts the 3-D offset with the feature pyramid for regression. 
As illustrated in Fig.~\ref{fig:pipeline}, we add one extra 
\textsc{Conv$_{3\times 3}$(ReLU(Conv$_{1\times1}$))} layer at different FPN levels and estimate an offset vector $\mathbf{d}=(d_0, d_1, d_2)\in\mathbb{R}^{3}$ for each prediction. $\mathbf{P}$ is then re-ordered using the predicted offsets in two steps
\begin{align}
    P'(i, j, l) &\leftarrow P(i +d_0, j + d_1, l)  \label{eq:offset1}\\
    P' (i, j, l) &\leftarrow P'(i', j', l + d_2) \label{eq:offset2},
\end{align}
where Eq.~\ref{eq:offset1} is to conduct feature offset in a 2-D space and Eq.~\ref{eq:offset2} is the offset across different scales. In Eq.~\ref{eq:offset2}, $i'$ and $j'$ are the rescaled coordinates of $i$ and $j$ at different FPN levels. Eq.~\ref{eq:offset1} is realized by a bilinear interpolation, and Eq.~\ref{eq:offset2} is conducted by a resizing of $P'(:, :, l + \lfloor d_2\rfloor  + 1)$ followed by a weighted average with $P'(:, :, l + \lfloor d_2\rfloor)$ for a decimal number $d_2$, where $\lfloor \cdot \rfloor$ is the floor function. Notably, the extra \textsc{conv} layers increase the computational cost slightly ($\sim 1\%$), but significantly improve the performance.
\subsection{Thresholding with Gaussian Mixture Model}

Previous works \cite{sohn2020simple,liu2021unbiased} require a static hyperparameter $\tau$ for pseudo-bboxes filtering. It fails to take into account that the model's prediction confidence varies across categories and iterations, which makes inconsistent targets and has a profound effect on performance~\cite{chen2022dense}. Furthermore, tuning the threshold on different datasets is tedious. 

Our goal is to find a way to automatically distinguish the positive from negative pseudo-bboxes. 
Specifically, we hypothesize that the score prediction $s^c$ for category $c$ is sampled from a Gaussian mixture (GMM) distribution $ \mathcal P(s^c)$ on all unlabeled data with two modalities, positive and negative. (see the score distribution in the subfigure of Fig. \ref{fig:pipeline})
\begin{equation}
\mathcal P(s^c)= w^c_n \mathcal{N}(s^c\vert \mu_n^c, (\sigma^c_n)^2)+w^c_p\mathcal{N}(s^c\vert  \mu_p^c, (\sigma^c_p)^2),
\end{equation}
where $\mathcal{N}(\mu, \sigma^2)$ denotes a Gaussian distribution, $w^c_n, \mu^c_n, (\sigma^c_n)^2$ and $w^c_p, \mu^c_p, (\sigma^c_p)^2$ represent the weight, mean and variance of negative and positive modalities, respectively. 
The Expectation-Maximization (EM) algorithm is then used to infer the posterior $\mathcal P(pos \vert s^c,  \mu_p^c, (\sigma^c_p)^2)$ which is the probability that detection should be set as the pseudo-target for the student, and the adaptive score threshold is determined as
\begin{equation}
    \tau^c = \mathop{\mathrm{argmax}}\limits_{s^c} \mathcal P(pos \vert s^c,  \mu_p^c, (\sigma^c_p)^2)
\end{equation}
In practice, we maintain a prediction queue of size $N$ ($N\sim 100$) for each class to fit GMM. Considering that the score distribution from a single-stage detector is strongly imbalanced as the majority of prediction is negative,  only the top $K=\sum_k(s_k)$ number of predictions are stored in a queue. The EM algorithm only accounts for $\sim 10\%$ training time increase. The threshold can then be adaptively determined \emph{w.r.t.} the model's performance at different training stages.
\section{Experiments}

In this section, we first evaluate
our solution on a series of SSOD benchmarks and then validate the effectiveness of each component through extensive ablation studies.\\
\textbf{Datasets and Evaluation Setup.} we conduct comprehensive experiment on the MS-COCO 2017~\cite{lin2014microsoft} benchmark and PASCAL VOC datasets~\cite{pascal-voc-2012}. 

\begin{figure}
\vspace{-2mm}
    \includegraphics[width=\linewidth]{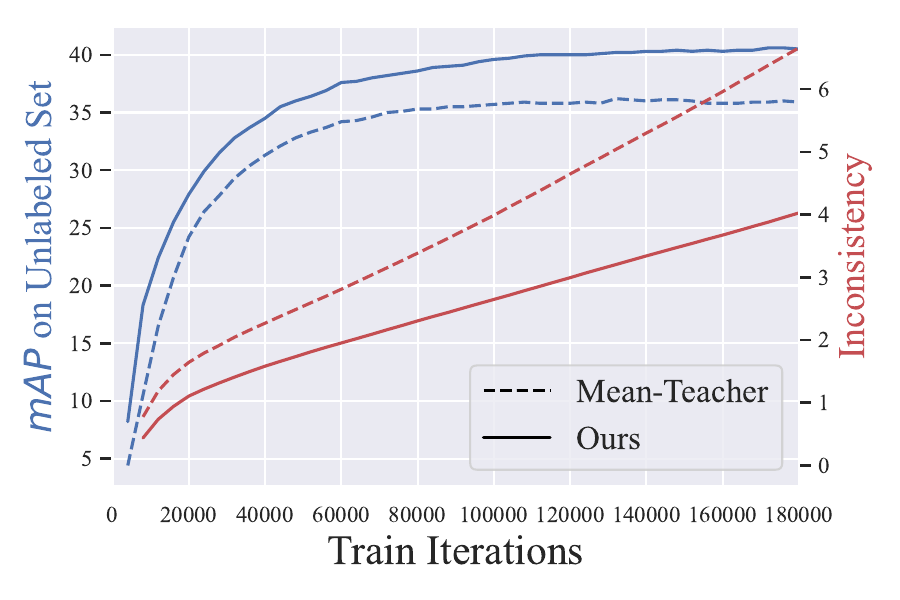}
    \vspace{-10mm}
    \caption{\name improves the training consistency in SSOD. (Left axis) mAP on the unlabeled set at different times. (Right axis) The inconsistency of pseudo labels.}
    \label{fig:consistency}
    \vspace{-4mm}
\end{figure}
\begin{figure*}[h]
    \centering
    \begin{minipage}{.32\linewidth}
    \vspace{-3mm}
    \centering
    \includegraphics[width=0.95\linewidth]{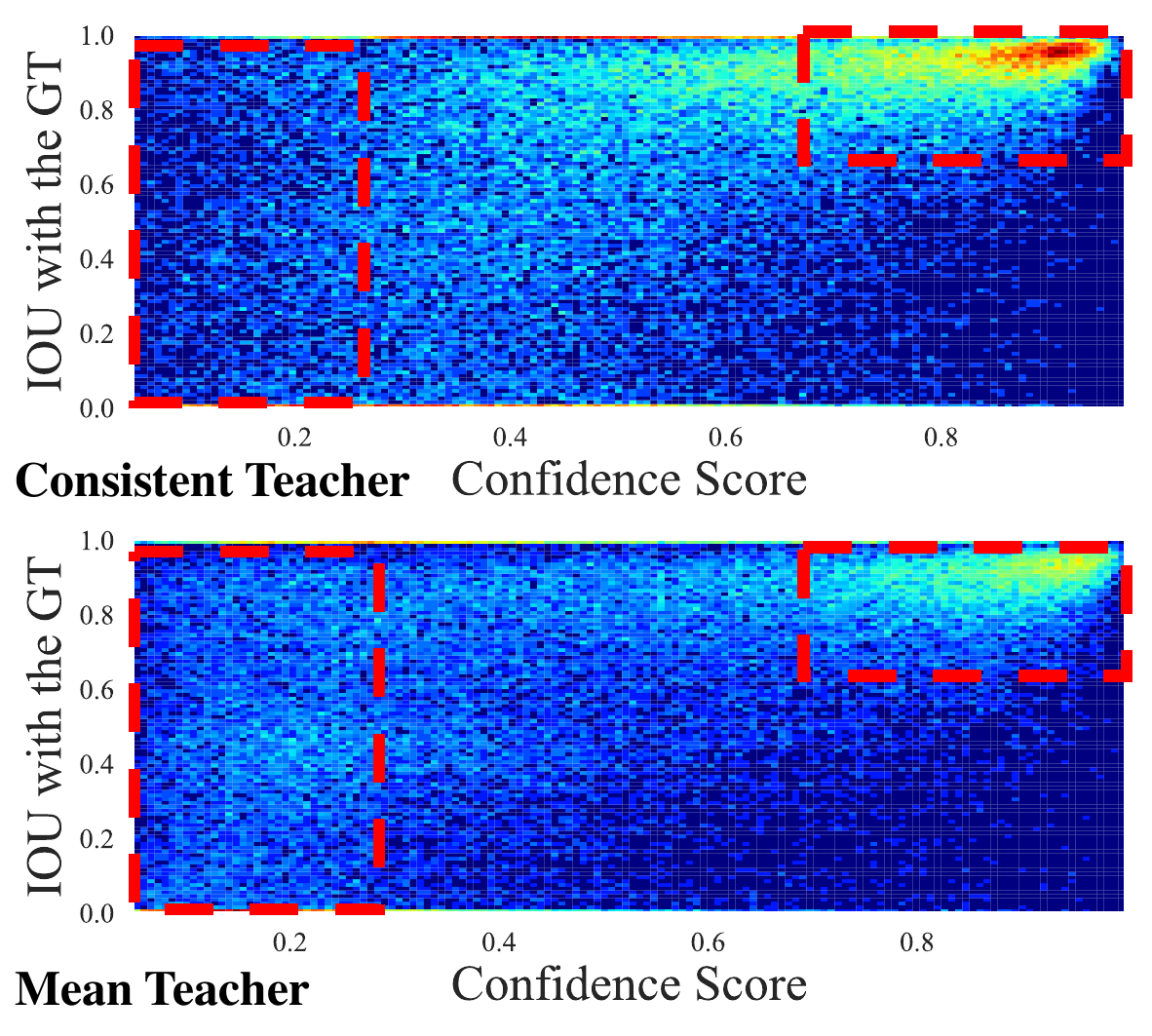}
    \vspace{-1mm}
    \caption{Heatmap of predicted bboxes confidence and its IoU score with GTs.}
    \label{fig:conf_IoU}
    \end{minipage}
    \begin{minipage}{.33\linewidth}
    \vspace{-1mm}
    \centering
    \includegraphics[width=\linewidth]{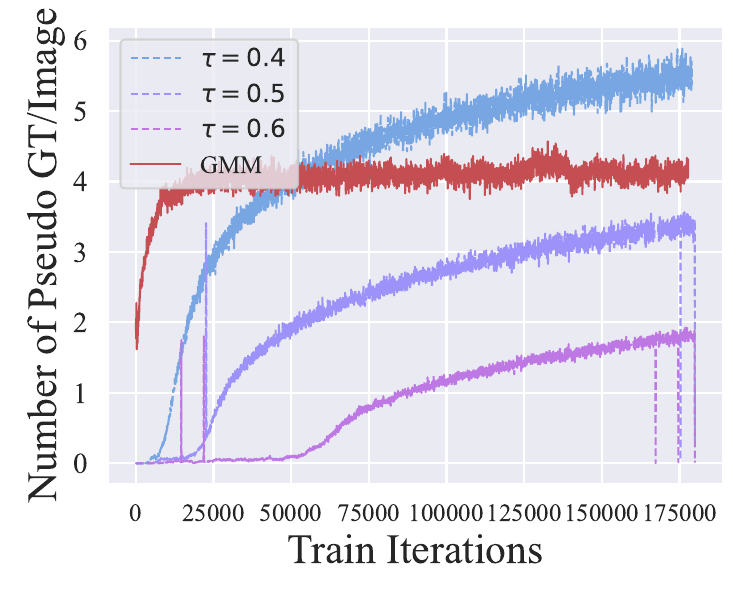}
    \vspace{-6mm}
    \caption{Number of pseudo labels/image with threshold schedules on COCO 10\%. }
    \label{fig:num_gt}
    \end{minipage}
    \hfill
    \begin{minipage}{.32\linewidth}
    \centering
    \includegraphics[width=1.03\linewidth]{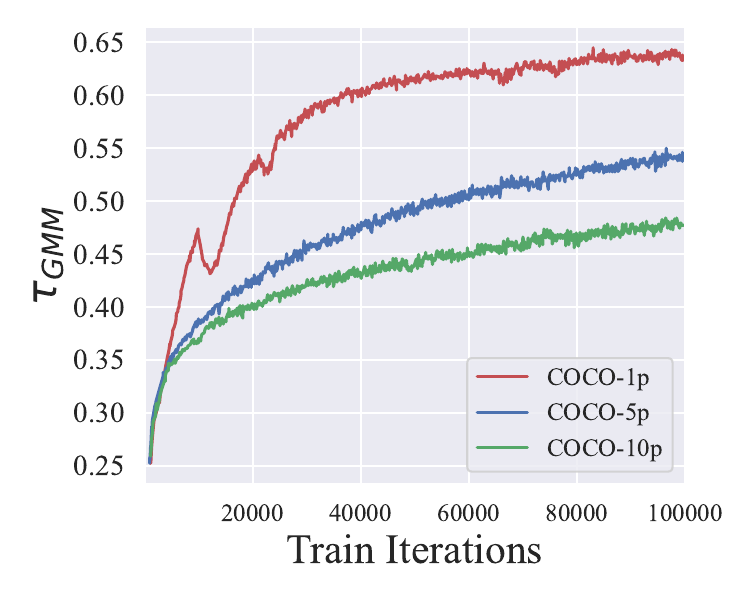}
    \vspace{-7mm}
    \caption{Average GMM thresholds across different classes  along with the training.}
    \label{fig:gmm_confidence}
    \end{minipage}
    \hfill
\vspace{-2mm}
\end{figure*}
We include three evaluation protocols: (1) \textsc{COCO-partial}: We randomly sample 1\%/2\%/5\%/10\% of the images in  \texttt{train2017} as labeled data and treat the rest as unlabeled data. We report the $AP_{50:95}$~\footnote{$AP_{50:95}$ is interchangable with mAP in this study.} results on the \texttt{val2017} as the evaluation metrics. (2) \textsc{COCO-addition}: We use the full \texttt{train2017} as labeled set and include the official unlabeled set \texttt{unlabel2017} as unlabeled set. The trained models are evaluated on \texttt{val2017}. (3) \textsc{VOC-partial}: We utilize the \texttt{VOC2007} trainval set as the labeled data and make use of the \texttt{VOC2012} trainval as our unlabeled data. The final model is verified on \texttt{VOC2007} test set using both $AP_{50}$ and $AP_{50:95}$ following~\cite{sohn2020simple}. Additionally, we evaluate the model improvements on the standard fully-supervised COCO-1x training~\cite{lin2017focal} to compare the relative benefits of our proposed method on both semi- and fully-supervised regimes.\\
\textbf{Implementation Details.} To ensure a fair comparison, all detectors are trained on 8 GPUs with 5 images per GPU~(1 labeled and 4 unlabeled images) similar to~\cite{xu2021end}. The detectors are optimized using SGD with a constant learning rate of $0.01$, a momentum of $0.9$, and a weight decay of $0.0001$. The unlabeled data weight is $\lambda_{U}=2$. No learning rate decay is applied. In \textsc{COCO-Partial} and \textsc{VOC-partial} evaluation, we train the detectors for 180K iterations, whereas we increase the training time on \textsc{COCO-addition} to 720K for better convergence. The teacher model is updated through EMA with a momentum of 0.9995. We follow the same data prepossessing and augmentation pipeline in~\cite{xu2021end}. We adopt RetinaNet~\cite{lin2017focal} with ResNet-50~\cite{lin2017feature} backbone as our baseline.  ImageNet~\cite{deng2009imagenet}-pretrained model is used as initialization. 

We compare our \name with numerous prevailing SSOD approaches including CSD~\cite{jeong2019consistency}, STAC~\cite{sohn2020simple}, Instant Teaching~\cite{zhou2021instant}, Humble Teacher~\cite{tang2021humble}, Unbiased Teacher v1 and v2~\cite{liu2021unbiased,liu2022unbiased}, Soft Teacher~\cite{xu2021end}, ACRST~\cite{zhang2021semi}, DSL~\cite{chen2022dense}, S4OD~\cite{zhang2022s4od}, Dense Teacher~\cite{zhou2022dense} and PseCo~\cite{li2022pseco}. In addition, we implement a baseline method where students are trained using labeled and pseudo-labeled data, and the teacher is updated through a moving average of the student. We name it the Mean-Teacher baseline~\cite{tarvainen2017mean}. The default confidence threshold is set as 0.4. 

\subsection{Troubleshooting the Inconsistencies in SSOD}
At first, we provide a thorough analysis to justify inconsistencies in SSOD, and how our solution addresses them.

\textbf{Inconsistency Leading to Noisy Labels.} We plot the mAP of the pseudo-bboxes against the GT targets on unlabeled data in Figure~\ref{fig:consistency}(Left axis). It stands for the quality of the labels. In addition, the \emph{inconsistency} is measured, which is an accumulation of the mismatch between the pseudo-bboxes of two consecutive teacher checkpoints (Right axis). Please refer to the supplementary for the full formulation.

According to Figure~\ref{fig:consistency}(Right axis), while the Mean-Teacher suffers large unfavorable inconsistencies during training, \name significantly reduces the target discrepancy at different time steps. Consequently, our model enjoys continuous improvement over time, and therefore provides high-quality labels for its student, as shown in Figure~\ref{fig:consistency}(Left axis).

\textbf{Inconsistency Caused by Classification-Regression Misalignment.} It is a well-known problem in object detection that, the classification score may not fully reflect the regression quality~\cite{xu2021end,zhang2020bridging}. It deters the essence of SSOD since we rely heavily on the prediction score to filter labels. Figure~\ref{fig:conf_IoU} visualizes the confidence-IoU heatmap of all predicted bounding boxes on the COCO \texttt{val2017}. For each predicted bbox, we plot the confidence of the maximum category and its maximum IoU with the GT boxes in the corresponding class. As highlighted in the red squares, Mean-Teacher predicts low-confidence but high-IoU bboxes. On the other hand, our model generates predictions that are concentrated in high-confidence and high-IoU regions.
\name gives rise to more calibrated predictions. 

A demo video is attached in the Supplementary Material to illustrate that cls-reg misalignment leads to shifting and noisy targets. Our FAM-3D largely prevents low-quality, but high-score noise predictions thus reducing inconsistency.

\textbf{Inconsistency Caused by Hard Score Threshold.} 
Figure~\ref{fig:num_gt} plots the number of pseudo GTs per image on the unlabeled data using different thresholding schedules. Notably,
it reveals a critical problem that, with static confidence thresholds $\tau=0.4,0.5,0.6$, the number of pseudo labels keeps going up as the detector becomes more confident. 
GMM-based approach, on the other hand, adaptively adjusts the best threshold according to the model capacity, with a nearly constant number of GTs, which reduces temporal inconsistency. 
\begin{table*}[t]
    \centering
    \renewcommand{\arraystretch}{0.9}
    \caption{\textsc{COCO-partial} comparison with other semi-supervised detector on \texttt{val2017}. The results for two-stage~(upper half) and single-stage~(lower half) detectors are listed separately. We also report the Faster-RCNN and RetinaNet performance trained on labeled data only. All models adopt ResNet50 with FPN as the backbone. We highlight the previous best record with \underline{underline}.}
    \setlength{\tabcolsep}{2mm}{
    \begin{tabular}{l|c|c|c|c}
    \hline
        Method & 1\% COCO& 2\% COCO& 5\% COCO &10\% COCO \\
        \hline
        Labeled Only  &9.05 & 12.70 & 18.47 & 23.86\\
        CSD  &10.51&13.93&18.63&22.46\\
        STAC  &13.97&18.25&24.38&28.64\\
        Instant Teaching  & 18.05 & 22.45&26.75 & 30.40\\
        Humble teacher &16.96&21.72&27.70&31.61\\
        Unbiased Teacher &20.75&24.30&28.27&31.50\\
        Soft Teacher&20.46& - & 30.74 & 34.04
        \\
        ACRST & \underline{26.07}& \underline{28.69}& 31.35&34.92\\
        PseCo &  22.43 &  27.77 & 32.50 & 36.06\\
        \hline
          Labeled Only  & 10.22 &  13.80 & 19.40 & 24.10\\
        Unbiased Teacher v2 & 22.71& 26.03&30.08& 32.61\\
        DSL & 22.03& 25.19& 30.87& 36.22\\
        Dense Teacher & 22.38& 27.20 & \underline{33.01} & \underline{37.13}\\
        S4OD   &  20.10 & - & 30.00 & 32.90 \\
         Mean-Teacher & 20.40 & 26.00 & 30.40 & 35.50 \\
        \name & \textbf{25.30} & \textbf{30.40}  & \textbf{36.10} & \textbf{40.00}
        \\\hline
    \end{tabular}
    }
    \label{tab:coco_part}
\end{table*}
In Figure~\ref{fig:gmm_confidence}, we plot the estimated threshold curve obtained by GMM on COCO 1\%/5\%/10\%. The value steadily increases as training proceeds. Furthermore, with fewer labeled samples, GMM sets a higher confidence threshold in accordance with more overfitting issues. Typical static threshold setting is incapable to address the inconsistency in learning targets, while GMM provides a gratifying solution.

\subsection{Semi-supervised Object Detection}
In this section, we compare our method with previous state-of-the-art work under \textsc{COCO-partial}, \textsc{VOC-partial}, and \textsc{COCO-addition} evaluation protocol. 

\noindent\textbf{\textsc{COCO-partial} Results.} Table~\ref{tab:coco_part} systematically compares the mAP of all aforementioned semi-supervised detectors trained with COCO 1\%/2\%/5\%/10\% labels. We first note that the simple Mean Teacher baseline with RetinaNet detector constitutes a strong method for SSOD. It achieves an mAP of 35.5 on COCO 10\% experiments without sophisticated data re-weighting strategy or pseudo-labeling selection methods.  More surprisingly, \name achieves a remarkable progress over current methods on 2\%/5\%/10\% experiments.  It scores 36.1 and 40.0 mAP on COCO 5\%/10\% data, largely surpassing the best-performed model Dense Teacher by $\sim 3.1$ and $\sim 3$ mAP. 

         

\noindent\textbf{\textsc{VOC-partial} Results.} In addition to the COCO evaluations, we compare our proposed model against other SSOD approaches on VOC0712 datasets in Table~\ref{tab:voc_part}. Again, we notice that our \name makes outstanding improvements over its counterparts. Our method shows an improvement of $2.2$ absolute mAP compared with the latest state-of-the-art~\cite{liu2022unbiased, chen2022dense}. 

\noindent\textbf{COCO-addition Results.} Now we would like to push our model to its limits by taking the full COCO train \texttt{train2017} as labeled data and additional \texttt{unlabel2017} as unlabeled data. 
As shown in Table~\ref{tab:addition}, in the case of \textsc{COCO-addition}, our model achieves 47.7 mAP, surpassing all previous state-of-the-art works. 
\begin{table}[]
\renewcommand{\arraystretch}{0.9}
    \centering
        \caption{\textsc{COCO-addition} experimental results on \texttt{val2017} with \texttt{unlabel2017} as unlabeled set. Note that 1$\times$ represents 90K training iterations, and N$\times$ represents N$\times$90K iterations.}
        \label{tab:addition}
    \begin{tabular}{l|c}
    \hline
        Method & $AP_{50:95}$ \\
        \hline
        CSD(3$\times$) & 40.20$\xrightarrow[]{\text{\textcolor{red}{-1.38}}}$38.82\\
      STAC(6$\times$) & 39.48$\xrightarrow[]{\text{\textcolor{red}{-0.27}}}$39.21\\
        Unbiased Teacher(3$\times$) & 40.20$\xrightarrow[]{\text{\textcolor{applegreen}{+1.10}}}$41.30\\ 
        ACRST(3$\times$)  & 40.20$\xrightarrow[]{\text{\textcolor{applegreen}{+2.59}}}$42.79\\
        Soft Teacher(16$\times$)  & 40.90$\xrightarrow[]{\text{\textcolor{applegreen}{+3.70}}}$44.50\\
        DSL(2$\times$) & 40.20$\xrightarrow[]{\text{\textcolor{applegreen}{+3.60}}}$43.80 \\
        PseCo(8$\times$) & 41.00$\xrightarrow[]{\text{\textcolor{applegreen}{+5.10}}}$46.10 \\
        Dense Teacher(8$\times$) & 41.24$\xrightarrow[]{\text{\textcolor{applegreen}{+4.88}}}$46.12 \\\hline
        \name (8$\times$)  & 40.50$\xrightarrow[]{\textbf{\textcolor{applegreen}{+7.20}}}$\textbf{47.70}\\
        \hline
    \end{tabular}
\end{table}

\begin{table}[]
\renewcommand{\arraystretch}{1.0}
    \centering
        \centering
    \caption{\textsc{VOC-partial} experimental results comparison with other semi-supervised detector on \texttt{VOC07} labeled and \texttt{VOC12} unlabeled set. }
    \label{tab:voc_part}
    \begin{tabular}{l|c|c}
    \hline
        Method  & $AP_{50}$ & $AP_{50:95}$\\\hline
        Labeled Only & 72.63 & 42.13\\
        CSD & 74.70 & - \\
        STAC & 77.45 & 44.64 \\
        ACRST & 78.16 & 50.12 \\
        Instant Teaching & 79.20 & 50.00 \\
        Humble Teacher & 80.94 & 53.04\\
        Unbiased Teacher & 77.37 & 48.69 \\
        Unbiased Teacher v2 & \underline{81.29} & \underline{56.87} \\
         
         \hline  
        Mean-Teacher & 77.02 & 53.61 \\
        \name & \textbf{81.00} & \textbf{59.00} \\
        \hline
    \end{tabular}
    \vspace{1mm}
    
         \caption{Comparisons between IoU-based and our adaptive anchor assignment on COCO.}\label{tab:assignment_ablation}
    \centering
    \begin{tabular}{l|l|l}
    \hline
        Assignment & $AP_{50:95}^{1\times}$ & $AP_{50:95}^{10\%}$ \\\hline
        IoU-based &  38.4 &  35.50 \\ \hline
        our ASA  & 40.1\tiny{\textcolor{applegreen}{(+1.7)}} & 38.50\tiny{\textcolor{applegreen}{(+3.0)}} \\
        \hline
    \end{tabular}
\end{table}
\begin{figure*}[t]
    \centering
    \begin{minipage}{.48\linewidth}
    \vspace{-5mm}
    \includegraphics[width=\linewidth]{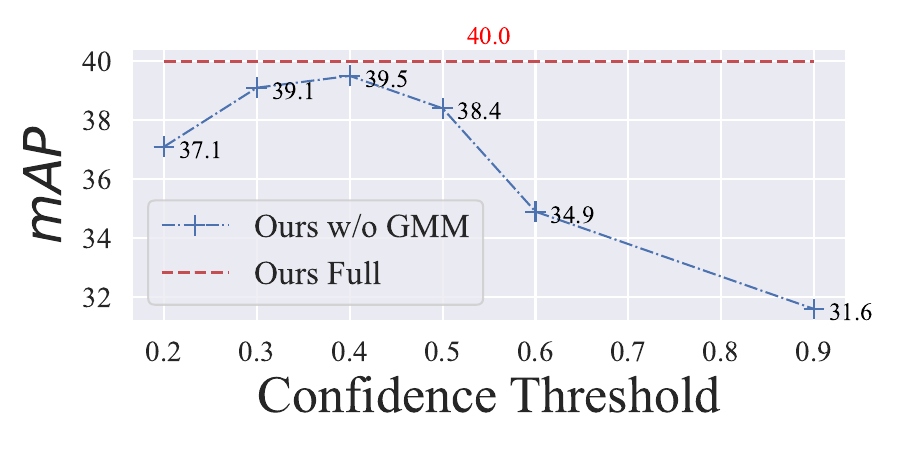}
    \vspace{-8mm}
        \caption{Ablative study of GMM-based pseudo-label filtering. Each value represents the mAP score on COCO 10\% data.}
    \label{fig:gmm}
    \end{minipage}
    \hfill
    \begin{minipage}{.48\linewidth}
    \vspace{-5mm}
    \includegraphics[width=\linewidth]{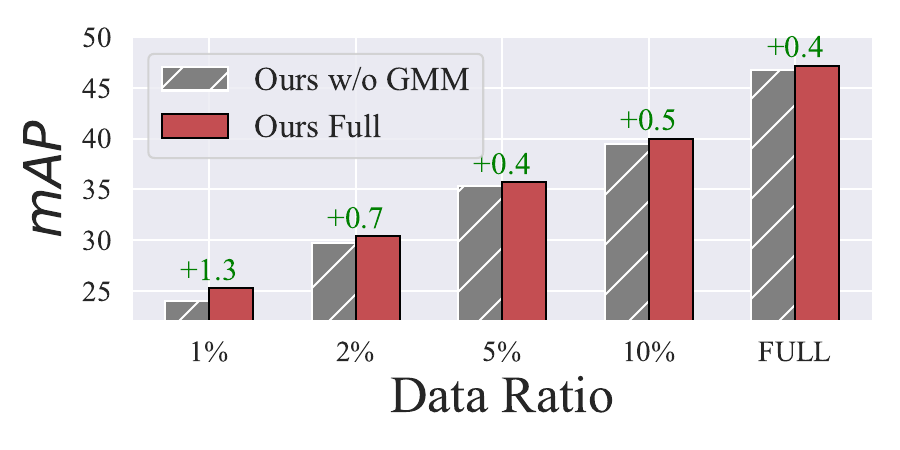}
    \vspace{-8mm}
        \caption{Ablation of GMM at different data ratio on COCO. Models are compared to baselines with a hard threshold 0.4.}
    \label{fig:gmm2}
    \end{minipage}
    \vspace{-2mm}
\end{figure*}
\subsection{Ablation Study}
In this section, we validate the effectiveness of our 3 major designs on the MS-COCO dataset.


\noindent\textbf{Adaptive Sample Assignment.} We first examine the effect of ASA strategy. To enable a fair comparison between all assigners, we utilize the Mean Teacher with a fixed confidence threshold of 0.4 and unlabeled weight of 2 as our baseline and replace its IoU-based assignment with our proposed ASA. Since the adaptive assignment is also applicable to the supervised scenario, we further experiment on the supervised MS-COCO with the standard 1$\times$~(12 epochs) training setting. It is notable that, as shown in Table~\ref{tab:assignment_ablation}, a robust sample assignment plays a pivotal role in SSOD. 
By specializing the assignment policy on semi-supervised tasks, our ASA achieves 38.50 mAP on COCO 10\%, with an improvement of 3 mAP compared with the heuristic matching cost using IoU. Another finding is that the performance benefit from ASA is almost doubled on SSOD (3.0 mAP) than on the fully supervised setting (1.7 mAP). It suggests our proposed ASA is particularly beneficial in the evaluation of the SSOD tasks, as also seen in Fig.~\ref{fig:motivation} of its ability to suppress the confirmation bias in SSOD. 


\noindent\textbf{3D Feature Alignment Module.} To testify to the effectiveness of FAM, we first replace the FAM-3D as a 2-D counterpart, which is adopted in~\cite{feng2021tood}. Table~\ref{tab:head_ablation} provides the ablative study of our method with different FAM structures. We observe that the FAM-3D surpasses the setting without feature alignment by 1.0 mAP and FAM-2D by 0.4 mAP on COCO 10\% evaluation, with 
negligible parameters and FLOPs. It is shown that, by automatically estimating the best 3D feature location for classification and regression, the semi-supervised detector is better calibrated to identify high-quality pseudo-labels. It is also noted that our FAM-3D brings much more gains under a semi-supervised setting than that in fully-supervised learning, validating its extra benefit in reducing the noises in SSOD.  
\renewcommand{\arraystretch}{0.9}
\begin{table}[]
    \centering
        \caption{Ablation Study on detection head structure. We compare the performance, model size, and FLOPs on different head structures on COCO 10\% and standard $1\times$ evaluation. FLOPs are measured on the input image size of $1280\times 800$.}
    \setlength{\tabcolsep}{0.2mm}{
    \begin{tabular}{l|l|l|l}
    \hline
        Method & FLOPs (G) & $AP_{50:95}^{1\times}$  & $AP_{50:95}^{10\%}$  \\
        \hline
        Ours w/o FAM & 205.21& 40.1 & 38.5 \\
        Ours w FAM-2D&  205.70&40.4\tiny{\textcolor{applegreen}{(+0.3)}} & 39.1\tiny{\textcolor{applegreen}{(+0.6)}} \\
        Ours w FAM-3D & 208.49&40.7\tiny{\textcolor{applegreen}{(+0.6)}} & 39.5\tiny{\textcolor{applegreen}{(+1.0)}} \\
        \hline
    \end{tabular}
    \vspace{-1mm}
    \label{tab:head_ablation}
    }
\end{table}


\noindent\textbf{GMM-based thresholding.} We testify to the detector's performance with or without the GMM-based pseudo-labeling. We replace it with a hard confidence threshold $\tau \in  (0.2,0.3,0.4,$ $0.5,0.6,0.9)$. 
Figure~\ref{fig:gmm} illustrates the test mAP on \texttt{val2017}. Notice that the detector is highly sensitive to the confidence threshold, with the optimal constant threshold at 0.4. By fitting the distribution of confidence, GMM dynamically adjusts the threshold for selecting pseudo-labels. This not only frees us from the tedious threshold tuning process but also allows 
for a gained accuracy and stabler supervision signal than a fixed threshold, achieving the final performance of 40.00 mAP with 0.5 mAP improvement on 10\% labeled data. GMM is also higher than the model using a hard threshold (0.4) at different ratios of labeled data as well, as illustrated in Figure~\ref{fig:gmm2}.


    


\section{Limitations and Future Work}
Despite the effectiveness of \name, it is currently mainly developed on traditional single-stage detectors. Its application to two-stage detectors and recent DETR-based~\cite{carion2020end} detectors is to be verified. Moreover, semi-supervised learning with pseudo-labels can accumulate errors due to inaccurate priors and human heuristics during the self-recurrent process. Our adaptive sample assignment strategy has replaced some human heuristics, such as anchor-based assignments, resulting in additional benefits for SSOD. It is believed that exploring more end-to-end approaches to semi-supervised learning could also bring similar advantages, which is an avenue for future research.


\section{Conclusion}
This paper offers a systematic investigation of the inconsistency issues that arise in SSOD, and proposes a straightforward yet effective semi-supervised object detector called \name as a solution. The proposed method employs adaptive anchor assignment, which identifies the positive anchor with the lowest matching costs, and FAM, which aligns classification and regression tasks by regressing the 3-D feature pyramid offsets. To address the threshold inconsistency problem in pseudo-bboxes, GMM is utilized to dynamically adjust the threshold for self-training. By integrating these three modules, our \name achieves a significant performance improvement over state-of-the-art methods on various SSOD benchmarks, demonstrating robust anchor assignment and consistent pseudo-bboxes.

{\small
\bibliographystyle{ieee_fullname}
\bibliography{egbib}
}

\newpage

In this supplementary material, we present more experimental quantitative results, a comparison of model sizes, and visualizations of bounding boxes, all of which serve to bolster the effectiveness of our proposed \name. Furthermore, we provide more details on our experimental methodology, implementation information, and hyper-parameter settings.  Our code is also attached for your reference.
\setcounter{section}{0}

\section{More details in \name}
\subsection{Inconsistency measurement.}
\label{sec:Inconsistency}
\textbf{Inconsistency} pertains to the problem of pseudo boxes being highly inaccurate and varying greatly at different stages of training. To address this issue, we measure the degree of variation in pseudo-bboxes across different training steps. Specifically, we achieve this by saving checkpoints every 4000 training steps and running inference on a subset of 5000 images from the unlabeled set using these checkpoints. The prediction output from the previous checkpoint is treated as the Ground Truth (GT), and we evaluate the Mean Average Precision (mAP) of the current checkpoint using the previous predictions as the reference. A higher mAP indicates more consistent pseudo targets. Then the inconsistency is measured by accumulating $1-mAP$ for these checkpoints to reflect the accumulated effect of noisy targets.  

\section{Verification of the Inconsistency in SSOD}

\noindent\textbf{Assignment Inconsistency under Noisy Pseudo Labels.} To illustrate that the conventional IoU-based or heuristic label assignment is problematic in SSOD, we intentionally inject random noise to the ground-truth bounding boxes and testify the assignment consistency by quantifying the assignment IoU~(A-IoU) of clean and noisy assignments. Suppose a bounding box $b=(x_1, y_1, x_2, y_2)$ is assigned to a set of $k$ anchors $A=\{a_1, \dots, a_k\}$. We add Gaussian noise to its coordinate with a noise ratio $\rho$, so that $b' = (x_1+\epsilon_{x_1} \times w, y_1 + \epsilon_{y_1} \times h, x_2+\epsilon_{x_2} \times w, y_2 + \epsilon_{y_2} \times h )$, in which $w$ and $h$ are width and height of the box. $\epsilon_{x_1}, \epsilon_{y_1}, \epsilon_{x_2}, \epsilon_{y_2}$ are sampled from a normal distribution $\mathcal{N}(0,\rho)$. The perturbed box $b'$ is matched to a new set of $l$ anchors $A'=\{a'_1, \dots, a'_l\}$. The A-IoU is computed as the intersection-of-union between $A$ and $A'$. The higher A-IoU score suggests the assignment is more robust to label noise. 


We evaluated the assignment consistency under two scenarios. Firstly, we calculated the assignment Intersection over Union (IoU) with varying degrees of noise ratio $\rho \in {0.1, 0.2, \dots, 0.5}$ using the final model. Secondly, we investigated how assignment consistency changes during training by reporting the Average-IoU (A-IoU) at different stages of training, with a constant $\rho$ value of 0.1. We compared our ASA with IoU-based assigners~\cite{ren2015faster, lin2017focal, liu2016ssd} and ATSS assigner~\cite{zhang2020bridging}, using the Mean Teacher RetinaNet baseline on COCO 10\%. To ensure a fair comparison, we kept all modules identical except for the assignment module. In both evaluations, we randomly selected 1000 images from \texttt{val2017} to compute the A-IoU.

\begin{figure}[t]
    \centering
    \includegraphics[width=0.8\linewidth]{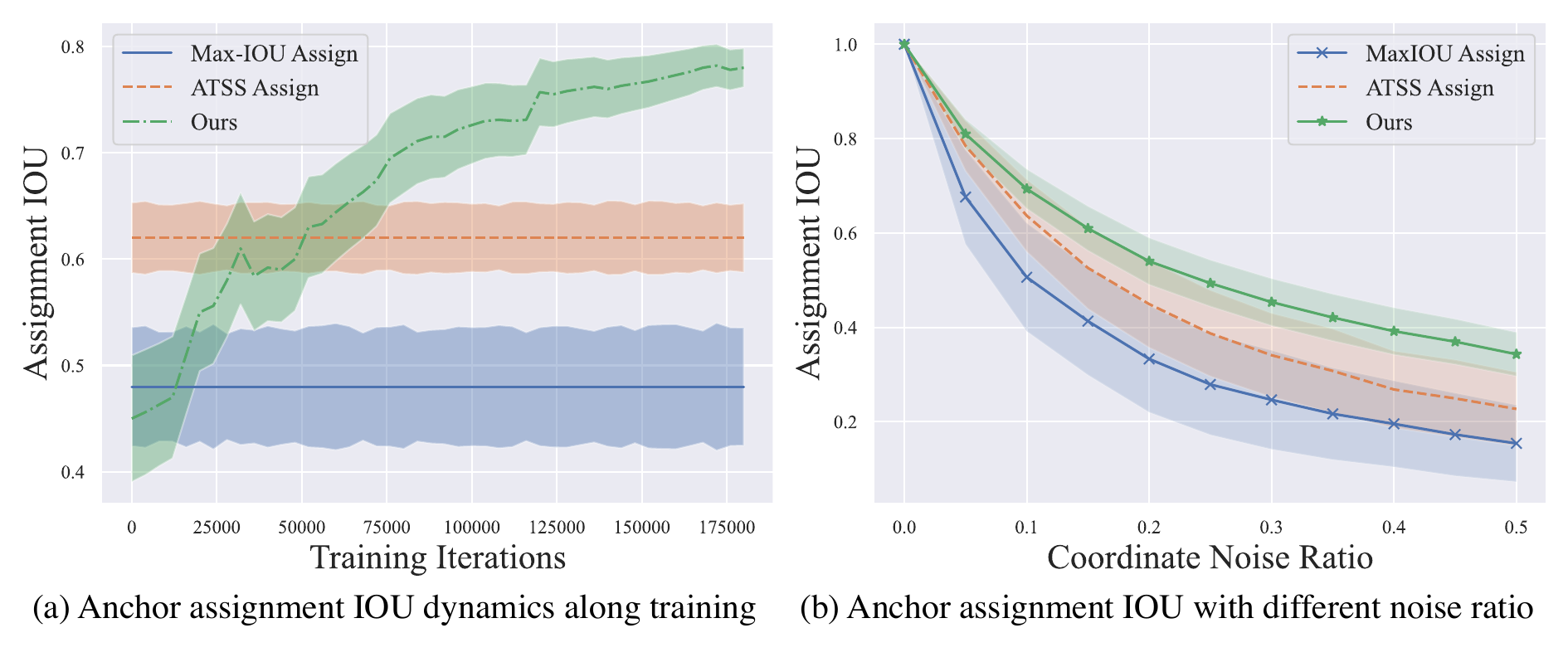}
    \caption{Assignment IoU score between ground-truth and the noisy bounding boxes (a) at different times of training and (b) using different noise ratios.}
    \label{fig:a-iou}
\end{figure}

Figure~\ref{fig:a-iou} depicts the mean$\pm$std A-IoU between clean and noisy labels at various training times and noise ratios $\rho$. In particular, Figure~\ref{fig:a-iou}(a) illustrates that both ATSS and our ASA achieve higher A-IoU than the commonly used IoU-based assignment. It is worth noting, however, that ATSS still relies on heuristic matching rules between labels and anchor boxes. In contrast, our ASA steadily improves as the detector becomes more accurate.

Figure~\ref{fig:a-iou}(b) illustrates that the IoU-based assignment method fails to maintain the initial assignment when a large amount of label noise is introduced. This experiment highlights that the IoU-based assignment method is incapable of maintaining consistent assignments in SSOD due to the inherently noisy nature of pseudo-labels. In contrast, our proposed ASA strategy performs well even under severe noise scenarios. This result supports our argument that our consistent assignment strategy is robust to label noise in SSOD.
{\renewcommand{\arraystretch}{2.0}
\begin{table}[]
    \centering
    \scriptsize
    \caption{Classification and Regression inconsistency analysis using IOU-Confidence linear regression~(LR) error. We also provide the Mean Teacher IoU-Confidence plot on the right.}
    \begin{tabular}{l|c|c}
    \hline
         & LR Standard Error &\multirow{3}{*}{\includegraphics[width=0.2\linewidth]{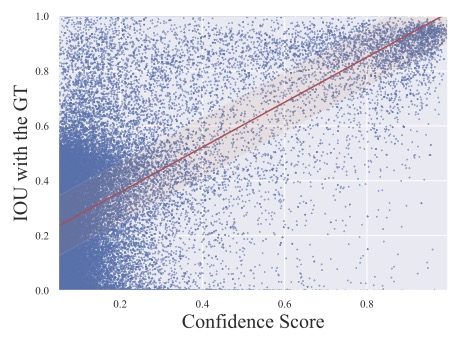}}
 \\\cmidrule{1-2}
       Mean Teacher  & 0.109 & \\
       \name  & \textbf{0.080} & \\
         \hline
    \end{tabular}
    \label{tab:iou_score}
\end{table}

}

\noindent\textbf{Classification and Regression Inconsistency.} We unveiled the regression and classification inconsistency problem by identifying the mismatch between the high-score and high-IoU predictions. We obtain the confidence-IoU pairs on \texttt{val2017} using \name and Mean Teacher RetinaNet when trained on COCO 10\% data, and analyze the correlation between the two variables. We apply linear regression and measure the standard error to reflect the correlation between confidences and IoUs. The smaller error indicates a higher correlation. 


Table~\ref{tab:iou_score} presents the linear regression~(LR) standard error for \name and Mean Teacher RetinaNet. The scatter plot on the right displays the confidence-IoU of Mean-Teacher. We observe a clear misalignment between classification and regression tasks in semi-supervised detectors, as numerous low-confidence predictions possess high IoU scores. This indicates that classification confidence does not provide a strong enough clue for accurate regression, resulting in erroneous pseudo-label noise during training. The high LR error of 0.109 with Mean Teacher RetinaNet further demonstrates this point.
In contrast, our \name largely eliminates the mismatch between the two tasks with a lower LR error of 0.080. This supports our argument that \name can align the classification and regression sub-tasks and reduce the mismatch in SSOD.

\section{Additional Ablation Study}
\subsection{Anchor-based VS Anchor-Free}
In this study, we aim to compare the performance of anchor-based and anchor-free object detectors on the MS-COCO 10\% SSOD benchmark dataset. To achieve this, we have selected RetinaNet as a representative anchor-based detector and FCOS as a representative anchor-free detector. We then apply the MeanTeacher baseline and our proposed \name, to see how different detectors perform on semi-supervised detection tasks. 

Table~\ref{tab:anchor} displays the performance of  both detectors, with and without the implementation of our proposed approach. The results demonstrate that our \name method substantially enhances the performance of both anchor-based and anchor-free baseline detectors. For instance, semi-supervised FCOS achieves a 35.8 mAP with MeanTeacher but experiences a 4.1 mAP increase when using our method. Additionally, the plug-and-play characteristic of our approach facilitates smooth integration with various detectors, underscoring its adaptability and effectiveness in augmenting object detection performance across distinct detector architectures.
\begin{table}[H]
    \centering
        \caption{SSOD performance with anchor-based and anchor-free detectors.}
    \label{tab:anchor}
    \begin{tabular}{l|c}
    \hline
         Method & mAP \\
          \hline
         FCOS MeanTeacher & 35.8 \\
         +\name & \textbf{39.9}\\
         \hline
         RetinaNet MeanTeacher & 35.5\\
         +\name & \textbf{40.0}\\
         \hline
    \end{tabular}

\end{table}

\subsection{Ablation on $\lambda_{dist}$}
In our experiments, $\lambda_{dist}$ is utilized to ensure stable training. However, in this section, we aim to investigate the impact of $\lambda_{dist}$ on the results. Specifically, we present the outcomes for various values of $\lambda_{dist}$, including ${0, 0.001, 0.002, 0.01}$, in Tab.~\ref{tab:lambda-dist}. Setting $\lambda_{dist}=0$ leads to highly unstable assignment, which can cause memory overflow, particularly during the initial phase of training when matching is quite inaccurate. On the other hand, when $\lambda_{dist}$ is significant, the centerness prior cancels out the performance advantage of our ASA. It is safe to set $\lambda_{reg}$ in ASA to the same value as that in the loss term. 
\begin{table}[]
 \caption{Ablation for the $\lambda_{dist}.$}
     \label{tab:lambda-dist}
     \centering
     \begin{tabular}{l|c|c|c|c}
     \hline
         $\lambda_{dist}$ & 0 & 0.001 & 0.002 &0.01  \\
         \hline
          mAP & Unstable & 40.0& 39.8	& 39.4\\
          \hline
     \end{tabular}
 \end{table}
\subsection{Training Time}
Table~\ref{tab:anchor} showcases the results comparing the training time per iteration for the RetinaNet-MeanTeacher detector on the MS-COCO SSOD task, employing various enhancements and methods. The impact of each method on the training time per iteration is evident from the table.

The RetinaNet baseline exhibits a training time of 1.25 s/iter. Intriguingly, ASA not only boosts performance but also reduces time complexity during the assignment, primarily due to its more efficient implementation and fewer anchor number requirement.  

FAM3D introduces a marginal increase in training time, suggesting a reasonable balance between performance enhancement and computational efficiency. In the case of GMM-based thresholding, updating the threshold every iteration results in an approximate 10\% increase in training time, indicating that GMM may provide certain advantages but at the cost of extended training durations.

\begin{table}[H]
 \centering
 \caption{Train time per second with different methods.}
    \label{tab:anchor}
\begin{tabular}{l|l|c}
\hline
Method & Sec./Iter. & $\Delta$\\
\hline
Improved RetinaNet & 1.25   &-         \\
+ ASA              & 1.18     &\textcolor{Green}{-0.07}      \\
+ FAM2D            & 1.22   & \textcolor{Red}{+0.04}       \\
+ FAM3D            & 1.26 & \textcolor{Red}{+0.04}\\
+ GMM              & 1.38 &\textcolor{Red}{+0.12}\\\hline
\end{tabular}

\end{table}

\section{Detection results visualization}
\subsection{Qualitative comparison with the baseline.} 
\begin{figure*}
    \centering
    \includegraphics[width=\linewidth]{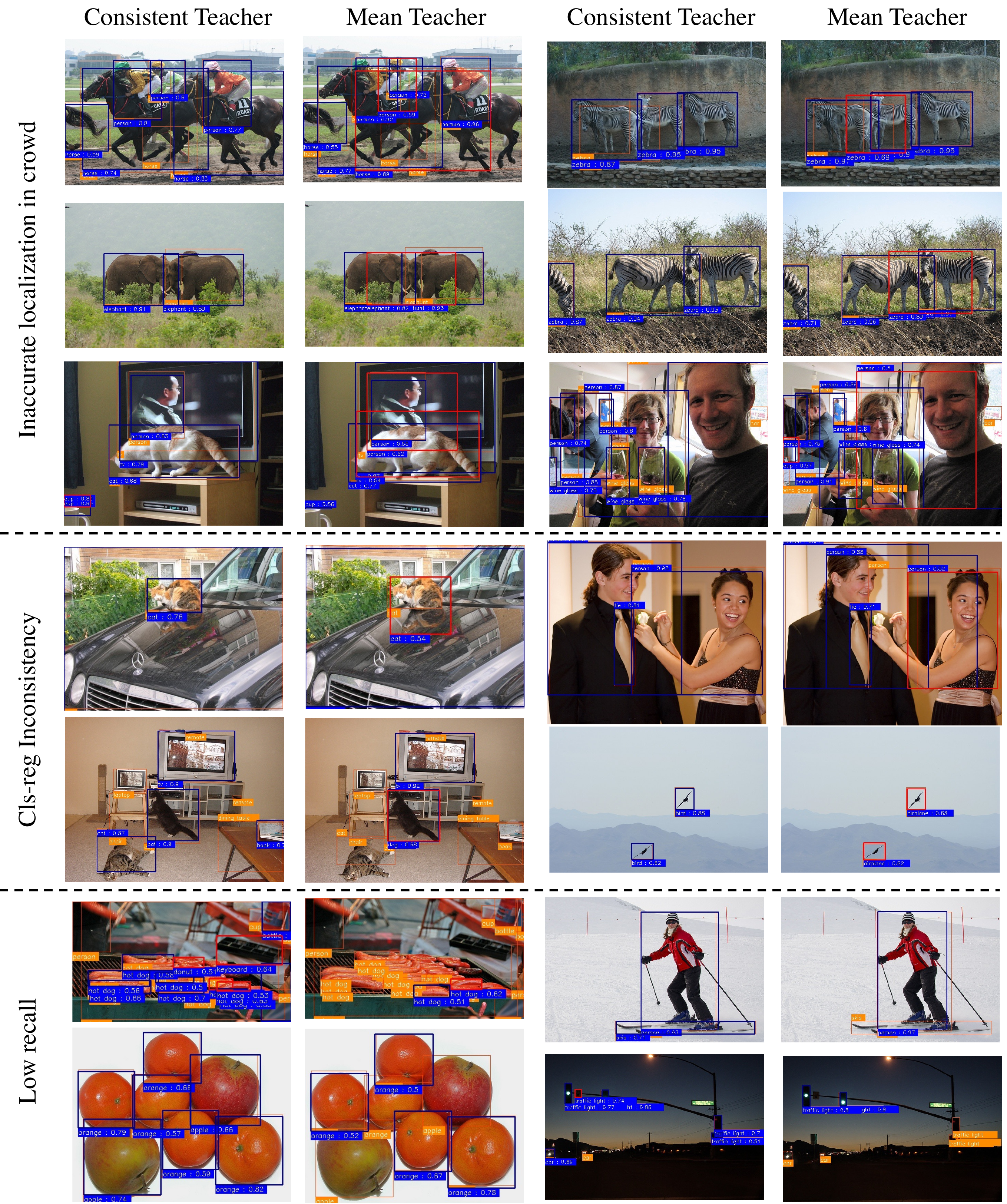}
    \caption{Qualitative comparison on the COCO\%10 evaluation. The bounding boxes in \textcolor{Orange}{Orange} are the ground truths, and \textcolor{Violet}{Violet}  refers to the prediction. \textcolor{Red}{Red} highlights the false positive predictions.}
    \label{fig:compare}
\end{figure*}

To further compare our \name with the baseline Mean Teacher RetinaNet, we visualize the predicted bounding boxes on \texttt{val2017} under the COCO 10\% protocol. In Figure~\ref{fig:compare}, we plot the predicted and ground-truth bounding boxes in \textcolor{Violet}{Violet} and \textcolor{Orange}{Orange} respectively, while highlighting the false positive bounding boxes in \textcolor{Red}{Red}.

There are 3 general properties that we could observe in our demonstration. 

\begin{enumerate}
\item Firstly, \name is better suited for crowded object localization than Mean Teacher. Mean Teacher often mistakes the intersection of two overlapped objects as a new instance, whereas \name largely resolves the inaccurate positioning problem through its adaptive anchor selection mechanism. For example, in scenes with zebras or sheep, Mean Teacher often gives a false positive output in the overlapping area of the two objects, whereas \name is able to accurately locate the objects.
\item Secondly, under the semi-supervised setting, Mean Teacher RetinaNet may either predict the wrong class for the correct location or regress an inaccurate bounding box despite having high classification confidence. For example, birds are sometimes misidentified as airplanes even when the localization is accurate. This is mainly due to the inconsistency between the classification and regression tasks, i.e., the features required for regression may not be optimal for classification. In contrast, \name effectively discriminates between similar categories using its FAM-3D module to dynamically select the most appropriate features.
\item Thirdly, \name achieves higher recall by being capable of detecting small or crowded instances that Mean Teacher may fail to identify. For example, \name is able to detect most of the hot dogs on a grill, while Mean Teacher may neglect most of them.
\end{enumerate}

\subsection{Good and Failure Cases.} 
\begin{figure*}
    \centering
    \includegraphics[width=\linewidth]{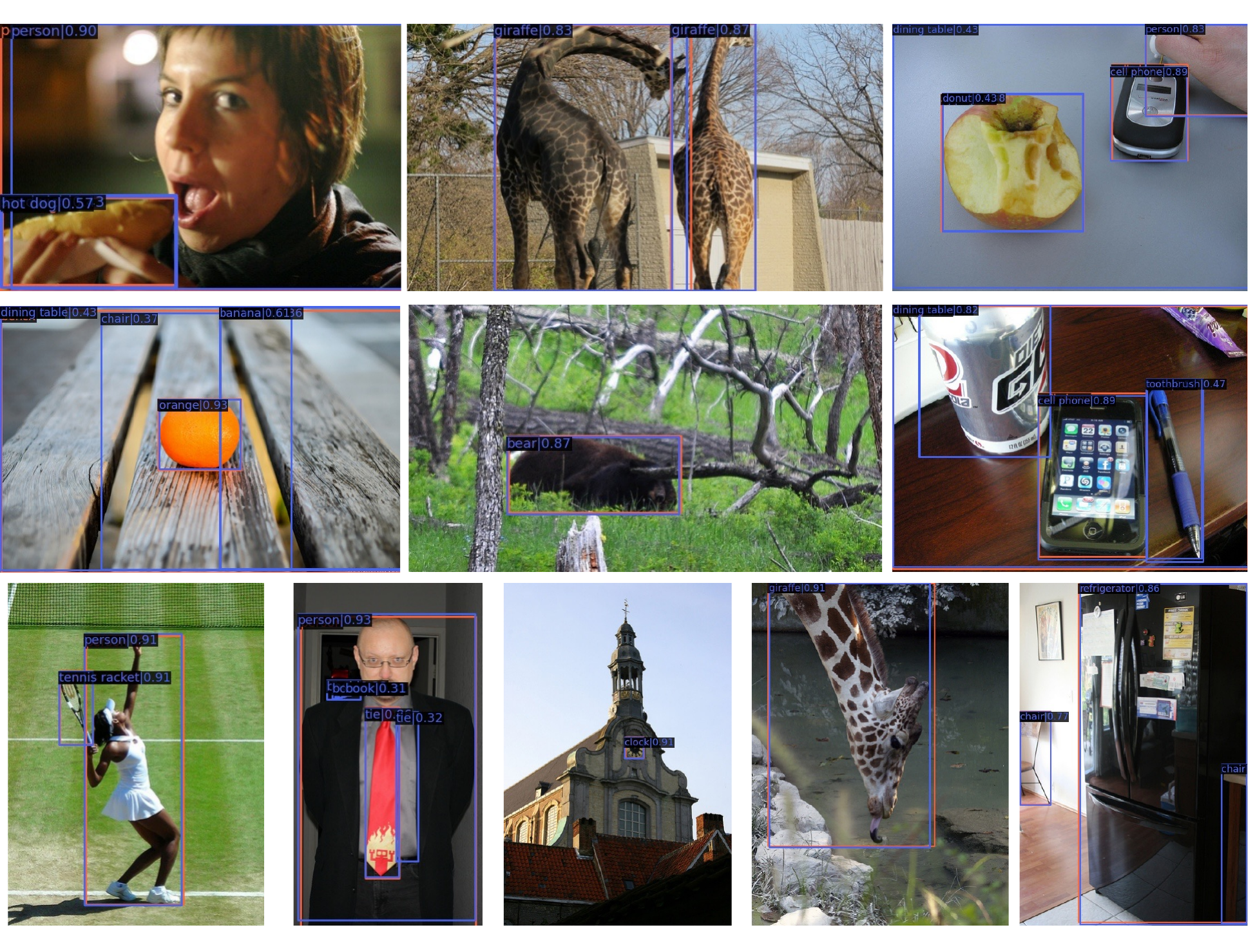}
    \caption{Good detection results for the COCO\%10 evaluation. The bounding boxes in \textcolor{Orange}{Orange} are the ground truths, and \textcolor{Violet}{Violet}  refers to the prediction.}
    \label{fig:good}
\end{figure*}

\begin{figure*}
    \centering
    \includegraphics[width=\linewidth]{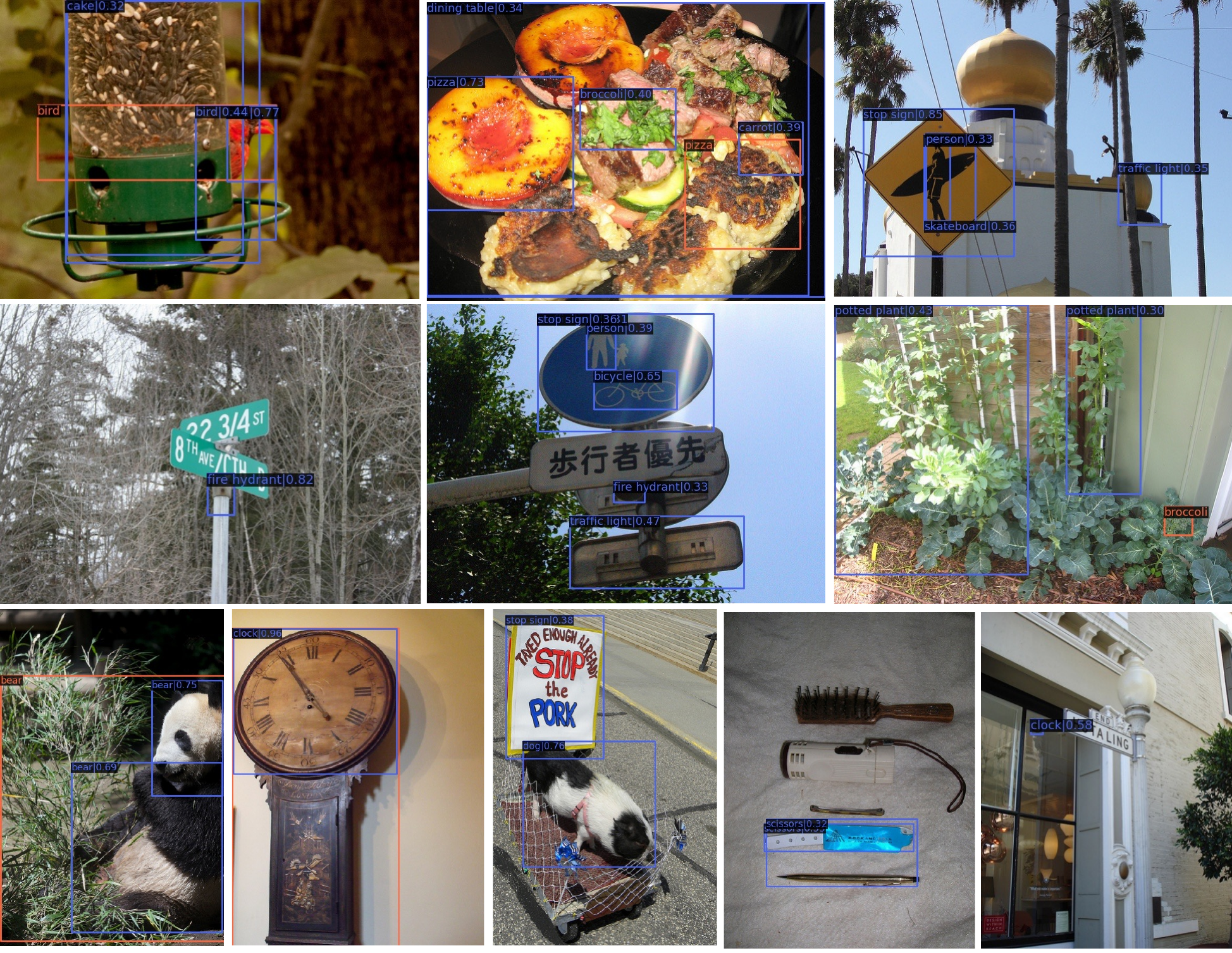}
    \caption{Failure detection results for the COCO\%10 evaluation. The bounding boxes in \textcolor{Orange}{Orange} are the ground truths, and \textcolor{Violet}{Violet}  refers to the prediction.}
    \label{fig:fail}
\end{figure*}

We provide additional examples to showcase the successful and unsuccessful instances produced by \name on COCO \texttt{val2017}, shown in Figure~\ref{fig:good} and Figure~\ref{fig:fail}, respectively. Although our proposed method has achieved impressive performance on a variety of SSOD benchmarks, Figure~\ref{fig:fail} highlights several deficiencies. Firstly, the trained detector lacks robustness to some out-of-distribution samples, such as cartoon characters on street signs being recognized as real people, and reflections in mirrors being identified as objects. Secondly, our detection performance is poor for some classes with small sizes, such as toothbrushes, hair dryers, etc. Thirdly, \name also tends to treat parts of the object as a whole, such as the head of a giant panda being detected as a separate animal~(in the lower left corner), and the dial of a clock being identified as the entire clock~(on the right of the panda).

\section{Experiment and Hyper-parameter settings}

\subsection{Datasets and Data Preprocessing}
\subsubsection{MS-COCO 2017} The Microsoft Common Objects in Context (MS-COCO) is a large-scale dataset used for object detection, segmentation, key-point detection, and captioning. In our SSOD experiments, we utilize the COCO2017 dataset, which includes 118K training and 5K validation images, along with bounding box annotations for 80 object categories.
\subsubsection{PASCAL VOC 2007-2012} The PASCAL Visual Object Classes (VOC) dataset contains 20 object categories, along with pixel-level segmentation annotations, bounding box annotations, and object class annotations. We adopt the official VOC 2007 \texttt{trainval} set, consisting of 5011 images, as the labeled set, and the 11540 images from the VOC 2012 \texttt{trainval} set as the unlabeled data in this study. Our evaluation is performed on the VOC 2007 test set.

\subsubsection{Data Augmentations.}
We use the same data augmentations as described in Soft Teacher~\cite{xu2021end}, including a labeled data augmentation in Table~\ref{tab:labeled}, a weak unlabeled augmentation in Table~\ref{tab:unlabeled weak} and a strong unlabeled augmentation in Table~\ref{tab:unlabeled strong}. 

\subsection{Implementation Details}
We implement our \name approach based on the MMdetection\footnote{https://github.com/open-mmlab/mmdetection} framework, using the data preprocessing code from the open-sourced Soft-Teacher\footnote{https://github.com/microsoft/SoftTeacher} and Google ssl-detection\footnote{https://github.com/google-research/ssl\_detection/}. We train our detectors on 8 NVIDIA Tesla V100 GPUs, and it takes approximately 3 days for 180K training iterations. Each GPU contains 1 labeled image and 4 unlabeled images. The source code is included in a separate zip file.

\newpage
\begin{table*}[]
    \centering
    \scriptsize
    \caption{Data augmentation for labeled image training.}
    \begin{tabular}{l|p{0.55\textwidth}|l}
    \hline
        Transformation & Description  &  Parameter Setting\\\hline
         RandomResize& Resize the image to the height of $h$ randomly sampled from  $h \sim U(h_{min}, h_{max})$,  while keeping the height-width ratio unchanged. & \pbox{3cm}{$h_{min} = 400, h_{max} = 1200$ in MS-COCO \\ $h_{min} = 480, h_{max} = 800$ in PASCAL-VOC} \\\hline
         RandomFlip&Randomly horizontally flip an image with a probability of $p$.& $p=0.5$ \\
         \hline
          OneOf & Select one of the transformations in a transformation set $T$.& $T=$  \texttt{TransAppearance}\\
         \hline
    \end{tabular}
    
    \label{tab:labeled}
\end{table*}
\begin{table*}[]
    \centering
    \scriptsize
    \caption{Weak data augmentation for an unlabeled image.}
    \begin{tabular}{l|p{0.55\textwidth}|l}
    \hline
        Transformation & Description  &  Parameter Setting\\\hline
         RandomResize& Resize the image to the height of $h$ randomly sampled from  $h \sim U(h_{min}, h_{max})$,  while keeping the height-width ratio unchanged. & \pbox{3cm}{$h_{min} = 400, h_{max} = 1200$ in MS-COCO \\ $h_{min} = 480, h_{max} = 800$ in PASCAL-VOC} \\\hline
         RandomFlip&Randomly horizontally flip an image with a probability of $p$.& $p=0.5$ \\
         \hline
    \end{tabular}
    \label{tab:unlabeled weak}
\end{table*}

\begin{table*}[]
    \centering
    \scriptsize
    \caption{Strong data augmentation for an unlabeled image.}
    \begin{tabular}{l|p{0.55\textwidth}|p{0.23\textwidth}}
    \hline
        Transformation & Description  &  Parameter Setting\\\hline
         RandomResize& Resize the image to the height of $h$ randomly sampled from  $h \sim U(h_{min}, h_{max})$,  while keeping the height-width ratio unchanged. & \pbox{3cm}{$h_{min} = 400, h_{max} = 1200$ in MS-COCO \\ $h_{min} = 480, h_{max} = 800$ in PASCAL-VOC} \\\hline
         RandomFlip&Randomly horizontally flip an image with a probability of $p$.& $p=0.5$ \\
         \hline
         OneOf & Select one of the transformations in a transformation set $T$.& $T=$  \texttt{TransAppearance}\\
         \hline
         OneOf & Select one of the transformation in a transformation set $T$.& $T=$  \texttt{TransGeo}\\
         \hline
         RandErase & Randomly selects $K$ rectangle region of size $\lambda h \times \lambda w$ in an image and erases its pixels with random values, where $(h,w)$ are the height and width of the original image. & \pbox{2cm}{$K \in U(1,5)$\\
         $ \lambda \in U(0,0.2)$}\\\hline
    \end{tabular}
    \label{tab:unlabeled strong}
\end{table*}

\begin{table*}[]
    \centering
       \scriptsize
    \caption{Appearance transformations, called \texttt{TransAppearance}.}
    \begin{tabular}{l|p{0.55\textwidth}|l}
    \hline
        Transformation & Description  &  Parameter Setting\\\hline
         Identity& Returns the original image.
         &\\\hline
        Autocontrast & Maximizes the image contrast by setting the darkest (lightest)
pixel to black (white). &\\\hline
Equalize & Equalizes the image histogram.& \\\hline
RandSolarize & Invert all pixels above a threshold value $T$.& $T \in U(0,1)$\\\hline
RandColor & Adjust the color balance of the image. $C=0$ returns a black\&white image, $C=1$ returns the original image. &$C \in U(0.05,0.95)$\\\hline
RandContrast & Adjust the contrast of the image. $C=0$ returns a solid grey image, $C=1$ returns the original image. &$C \in U(0.05,0.95)$\\\hline
RandBrightness & Adjust the brightness of the image. $C=0$ returns a black image, $C=1$ returns the original image. &$C \in U(0.05,0.95)$\\\hline
RandSharpness & Adjust the sharpness of the image. $C=0$ returns a blurred image, $C=1$ returns the original image. &$C \in U(0.05,0.95)$\\\hline
RandPolarize & Reduce each pixel to $C$ bits. &$C \in U(4,8)$\\\hline
    \end{tabular}
    \label{tab:TransAppearance}
\end{table*}

\begin{table*}[]
    \centering
       \scriptsize
    \caption{Geometric transformations, called \texttt{TransGeo}.}
    \begin{tabular}{l|p{0.55\textwidth}|l}
    \hline
        Transformation & Description  &  Parameter Setting\\\hline
        RandTranslate X& Translate the image horizontally by $\lambda\times$image width.
         & $\lambda \in U(-0.1,0.1)$\\\hline
         RandTranslate Y& Translate the image vertically by $\lambda\times$image height.
         & $\lambda \in U(-0.1,0.1)$\\\hline
          RandRotate Y& Rotates the image by $\theta$ degrees.
         & $\theta \in U(-30^\circ,30^\circ)$\\\hline
         RanShear X & Shears the image along the horizontal axis with rate $R$. & $R\in U(-0.480,0.480)$ \\\hline
         RanShear Y & Shears the image along the vertical axis with rate $R$. & $R\in U(-0.480,0.480)$ \\\hline
    \end{tabular}
    \label{tab:TransGeo}
\end{table*}

\clearpage
%
%

\end{document}


\pagestyle{headings}
\mainmatter
\def\ECCVSubNumber{2339}  

\title{Consistent Teacher Provides Better Supervision in Semi-supervised Object Detection\\-\textit{Supplementary Materials}-} 

\titlerunning{ECCV-22 submission ID \ECCVSubNumber} 
\authorrunning{ECCV-22 submission ID \ECCVSubNumber} 
\author{Anonymous ECCV submission}
\institute{Paper ID \ECCVSubNumber}

\maketitle

In this supplementary material, we provide additional experimental quantitative results, model size comparison, as well as bounding boxes visualization further to support the effectiveness of our proposed \name. In addition, we delineate more experimental details, implementation information, and hyper-parameter settings of our method. Our code is also attached for your reference.

\section{More details of \name}
\subsection{Balanced modality populations in Gaussian Mixture Model (GMM)}
Although GMM is able to identify the two modalities in the distribution of confidence scores, there is also an imbalance in the populations of the positive and negative samples. For example, there could be only one positive prediction, while the rest are all negative samples. The substantial imbalance not only harms the convergence speed using Expectation-Maximization (EM) algorithm but may also result in imprecise thresholds between positive and negative modalities. In practice, we only put top-$K$ predictions into the score bank used to estimate a GMM model, where $K=round(\sum_i^N b_i)$, where $b_i$ is the confidence score of a predicted bbox and $N \sim 100$ is the number of total predictions from the teacher network. In this way, the population between positive and negative samples is more balanced and the GMM estimated threshold becomes more accurate. 

\subsection{Loss function}
The loss function can be written as a general form
\begin{equation}
    \mathcal{L} = \mathcal{L}_s + \lambda_u(\mathcal{L}_u^{cls} + \lambda^{reg}\mathcal{L}_u^{reg}),
\end{equation}
where $\mathcal{L}_s$ and $\mathcal{L}_u$ denote supervised and unsupervised (unlabeled) losses, respectively.  $\lambda_u$ is the loss weight for $\mathcal{L}_u$ and $\lambda_{reg}$ is the loss weight for the regression subtask. In this study, GIoU loss~\cite{rezatofighi2019generalized} is used for regression, and parameters are set as $\mathcal{L}_u=2$ and $\lambda_{reg}=2$ across all different architectures. The standard Focal Loss~\cite{lin2017focal} is adopted for classification in the MeanTeacher baseline that consists of an EMA-updated RetinaNet model, while the Quality focal loss~\cite{li2021generalized} is used in \name for better alignment between classification and regression tasks.

\section{Semi-supervised detection results visualization}
\subsubsection{Qualitative comparison with Baseline} 
\begin{figure}
    \centering
    \includegraphics[width=\linewidth]{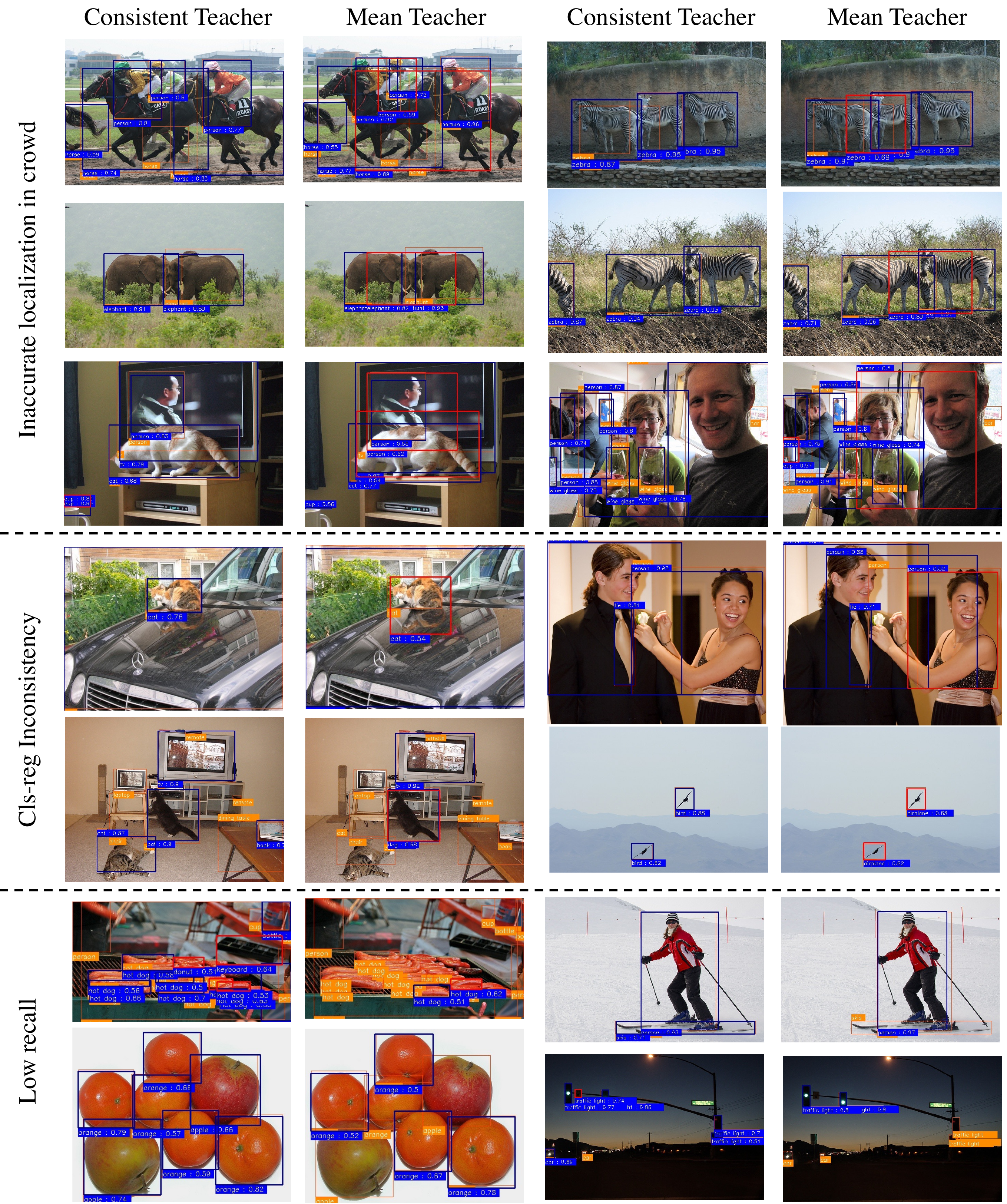}
    
    \caption{Qualitative comparison on the COCO\%10 evaluation. The bounding boxes in \textcolor{Orange}{Orange} is the ground-truth, and \textcolor{Violet}{Violet}  refers to the prediction. \textcolor{Red}{Red} highlights the false positive predictions.}
    \label{fig:compare}
\end{figure}

We further compare our \name with the baseline Mean Teacher RetinaNet by visualizing the predicted bounding boxes on \texttt{val2017} under the COCO 10\% protocol. In Figure~\ref{fig:compare}, we plot the predicted and ground-truth bounding boxes in \textcolor{Violet}{Violet} and \textcolor{Orange}{Orange} respectively, alongside with the false positive bboxes highlighted in \textcolor{Red}{Red}. 

There are 3 general properties that we could observed in our demonstration. 
\begin{enumerate}
    \item First, \name better fits the situation for crowded object localization, whereas Mean Teacher often mistakes the intersection of two overlapped objects as a new instance. For example, in 
scenes of zebras or sheep, Mean Teacher often gives a false positive output in the overlapping area of the two objects, where \name largely resolved the inaccurate positioning problem through the adaptive anchor selection mechanism.
\item On the other hand, we see that under the semi-supervised setting, the Mean Teacher RetinaNet could either predict the wrong class for the correct location or regress inaccurate bounding boxes when the classification confidence is high. For example,  birds are sometimes misidentified as airplanes even when the localization is accurate. It is mainly attributable to the inconsistency of classification and regression tasks, i.e. the features required for regression may not be optimal for classification. \name effectively discriminates similar categories using the FAM-3D to select the features dynamically.
\item Third, \name embraces higher recall since it is capable of detecting small or crowded instances where Mean Teacher fails to point out. For example, \name finds most of the hot dogs on the grill while Mean Teacher neglects most of them.
\end{enumerate}

\subsubsection{Good cases and Failure cases} 
\begin{figure}
    \centering
    \includegraphics[width=\linewidth]{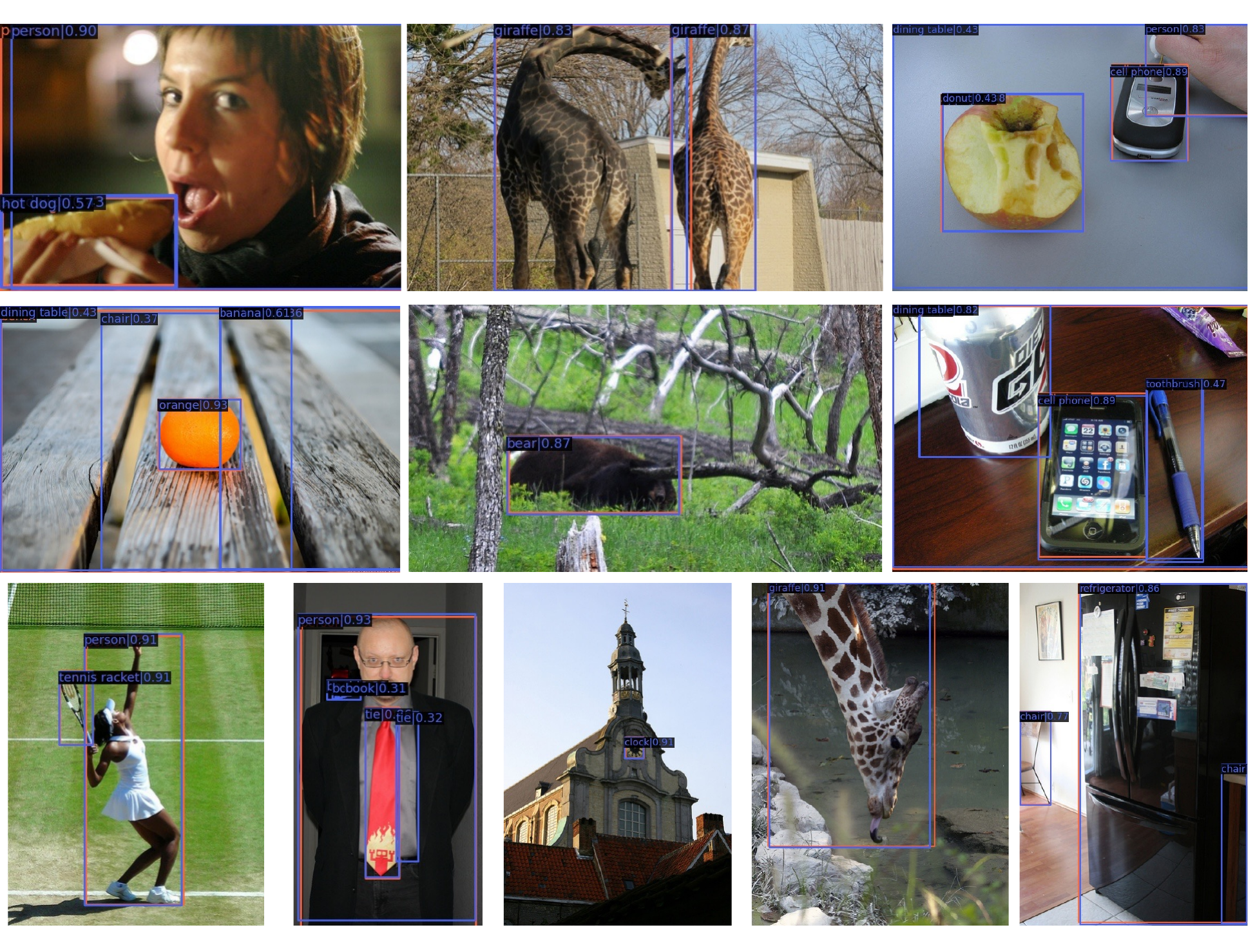}
    
    \caption{Good detection results for the COCO\%10 evaluation. The bounding boxes in \textcolor{Orange}{Orange} is the ground-truth, and \textcolor{Violet}{Violet}  refers to the prediction.}
    \label{fig:good}
\end{figure}

\begin{figure}
    \centering
    \includegraphics[width=\linewidth]{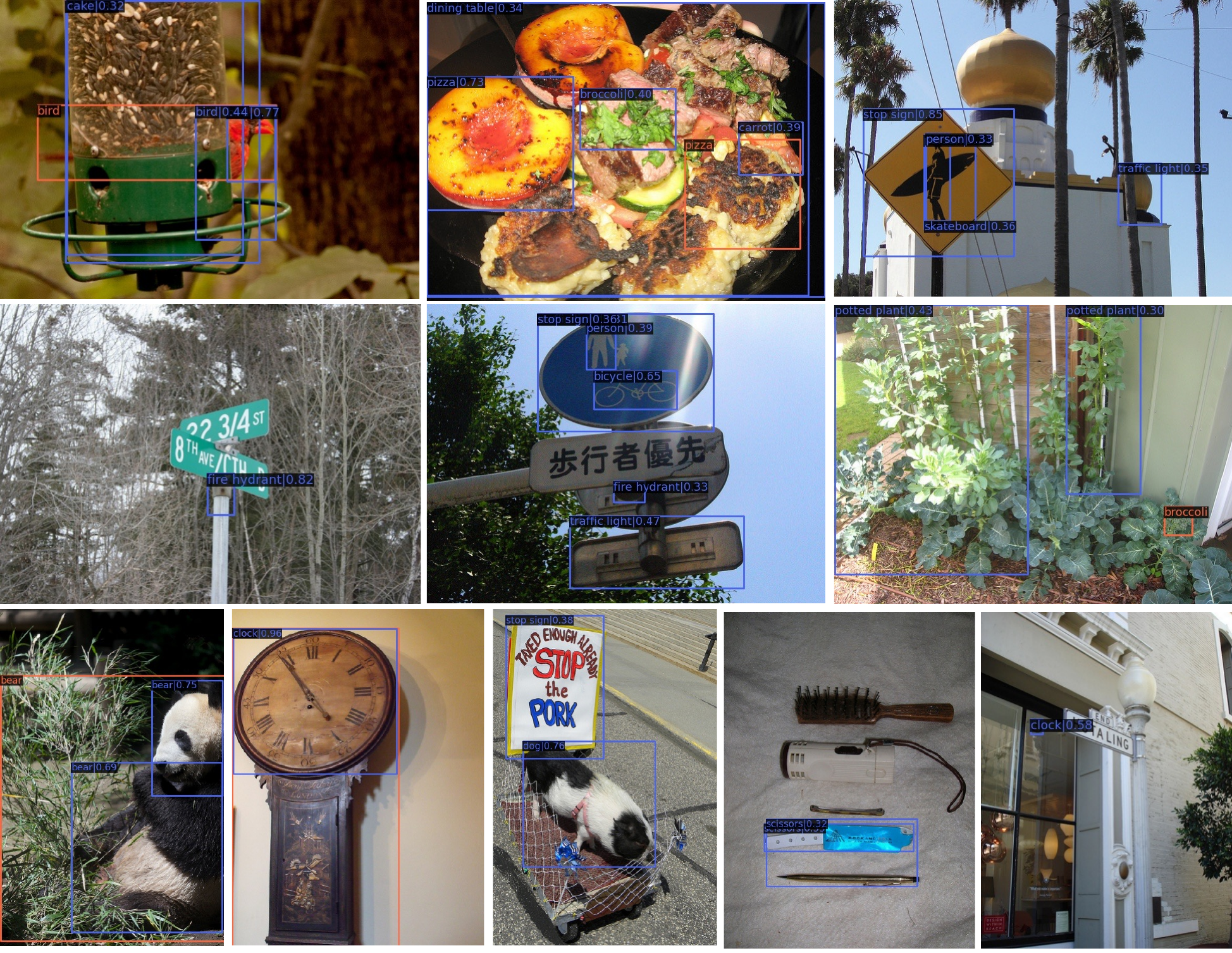}
    
    \caption{Failure detection results for the COCO\%10 evaluation. The bounding boxes in \textcolor{Orange}{Orange} is the ground-truth, and \textcolor{Violet}{Violet}  refers to the prediction.}
    \label{fig:fail}
\end{figure}

We provides more examples to showcase the good and failure examples produced by \name on COCO \texttt{val2017} in Figure~\ref{fig:good} and Figure~\ref{fig:fail}. Although our proposed method achieved gratifying performance on a series of SSOD benchmarks, we can still point out its deficiencies in Figure~\ref{fig:fail}. First, the trained detector lacks robustness to some out-of-distribution samples, for example, cartoon characters on street signs are recognized as real people, and reflections in mirrors are recognized as objects. Second, our detection performance is poor for some classes with small sample size, such as toothbrushes, hair dryers, etc. Third, it also tends to treat parts of the object as a whole, such as the head of the giant panda as a separate individual~(in the lower left corner), and the dial of a clock as the entire clock~(on the right of the panda).

\section{Experiment and Hyper-parameter settings}

\subsection{Datasets and data prepossessing.}
\subsubsection{MS-COCO 2017} The Microsoft Common Objects in Context~(MS-COCO) is a large-scale object detection, segmentation, key-point detection, and captioning dataset. We includes COCO2017 in our experiments for SSOD, which includes 118K/5K training validation images along with bounding boxes with 80 object categories. 
\subsubsection{PASCAL VOC 2007-2012.} The PASCAL Visual Object Classes (VOC) dataset contains 20 object categories alongside with pixel-level segmentation annotations, bounding box annotations, and object class annotations. The official VOC 2007 \texttt{trainval} set is adopted as the labeled set with 5011 images and the 11540 images from VOC 2012 \texttt{trainval} set is the unlabeled data. We evaluate on the VOC 2007 test set.

\subsubsection{Data Augmentations.}
We use the same data augmentations as described in Soft Teacher~\cite{xu2021end}, including a labeled data augmentation in Table~\ref{tab:labeled}, a weak unlabeled augmentation in Table~\ref{tab:unlabeled weak} and a strong unlabeled augmentation in Table~\ref{tab:unlabeled strong}. 

\subsection{Implementation Details} We implement our \name based on MMdetection\footnote{https://github.com/open-mmlab/mmdetection} framework, as the data prepossessing code borrowed from open-sourced Soft-Teacher~\footnote{https://github.com/microsoft/SoftTeacher} and google ssl-detection~\footnote{https://github.com/google-research/ssl\_detection/}. We train our detectors on 8 NVIDIA Tesla V100 GPUs. It takes approximately 3 days for a 180K training. Each GPU contains 1 labeled image a and 4 unlabeled images. The source code is attached in the a separate zip file.

\newpage
\begin{table}[]
    \centering
    \scriptsize
    \caption{Data augmentation for labeled image training.}
    \begin{tabular}{l|p{0.55\textwidth}|l}
    \hline
        Transformation & Description  &  Parameter Setting\\\hline
         RandomResize& Resize the image to a the height of $h$ randomly sampled from  $h \sim U(h_{min}, h_{max})$,  while keeping the height-width ratio unchanged. & \pbox{3cm}{$h_{min} = 400, h_{max} = 1200$ in MS-COCO \\ $h_{min} = 480, h_{max} = 800$ in PASCAL-VOC} \\\hline
         RandomFlip&Randomly horizontally flip a image with probability of $p$.& $p=0.5$ \\
         \hline
          OneOf & Select one of the transformation in a transformation set $T$.& $T=$  \texttt{TransAppearance}\\
         \hline
    \end{tabular}
    
    \label{tab:labeled}
\end{table}
\begin{table}[]
    \centering
    \scriptsize
    \caption{Weak data augmentation for unlabeled image.}
    \begin{tabular}{l|p{0.55\textwidth}|l}
    \hline
        Transformation & Description  &  Parameter Setting\\\hline
         RandomResize& Resize the image to a the height of $h$ randomly sampled from  $h \sim U(h_{min}, h_{max})$,  while keeping the height-width ratio unchanged. & \pbox{3cm}{$h_{min} = 400, h_{max} = 1200$ in MS-COCO \\ $h_{min} = 480, h_{max} = 800$ in PASCAL-VOC} \\\hline
         RandomFlip&Randomly horizontally flip a image with probability of $p$.& $p=0.5$ \\
         \hline
    \end{tabular}
    \label{tab:unlabeled weak}
\end{table}

\begin{table}[]
    \centering
    \scriptsize
    \caption{Strong data augmentation for unlabeled image.}
    \begin{tabular}{l|p{0.55\textwidth}|p{0.23\textwidth}}
    \hline
        Transformation & Description  &  Parameter Setting\\\hline
         RandomResize& Resize the image to a the height of $h$ randomly sampled from  $h \sim U(h_{min}, h_{max})$,  while keeping the height-width ratio unchanged. & \pbox{3cm}{$h_{min} = 400, h_{max} = 1200$ in MS-COCO \\ $h_{min} = 480, h_{max} = 800$ in PASCAL-VOC} \\\hline
         RandomFlip&Randomly horizontally flip a image with probability of $p$.& $p=0.5$ \\
         \hline
         OneOf & Select one of the transformation in a transformation set $T$.& $T=$  \texttt{TransAppearance}\\
         \hline
         OneOf & Select one of the transformation in a transformation set $T$.& $T=$  \texttt{TransGeo}\\
         \hline
         RandErase & Randomly selects $K$ rectangle region of size $\lambda h \times \lambda w$ in an image and erases its pixels with random values, where $(h,w)$ are height and width of the original image. & \pbox{2cm}{$K \in U(1,5)$\\
         $ \lambda \in U(0,0.2)$}\\\hline
    \end{tabular}
    \label{tab:unlabeled strong}
\end{table}

\begin{table}[]
    \centering
       \scriptsize
    \caption{Appearance transformations, called \texttt{TransAppearance}.}
    \begin{tabular}{l|p{0.55\textwidth}|l}
    \hline
        Transformation & Description  &  Parameter Setting\\\hline
         Identity& Returns the original image.
         &\\\hline
        Autocontrast & Maximizes the image contrast by setting the darkest (lightest)
pixel to black (white). &\\\hline
Equalize & Equalizes the image histogram.& \\\hline
RandSolarize & Invert all pixels above a threshold value $T$.& $T \in U(0,1)$\\\hline
RandColor & Adjust the color balance of image. $C=0$ returns a black\&white image, $C=1$ returns the original image. &$C \in U(0.05,0.95)$\\\hline
RandContrast & Adjust the contrast of image. $C=0$ returns a solid grey image, $C=1$ returns the original image. &$C \in U(0.05,0.95)$\\\hline
RandBrightness & Adjust the brightness of image. $C=0$ returns a black image, $C=1$ returns the original image. &$C \in U(0.05,0.95)$\\\hline
RandSharpness & Adjust the sharpness of image. $C=0$ returns a blurred image, $C=1$ returns the original image. &$C \in U(0.05,0.95)$\\\hline
RandPolarize & Reduce each pixel to $C$ bits. &$C \in U(4,8)$\\\hline
    \end{tabular}
    \label{tab:TransAppearance}
\end{table}

\begin{table}[]
    \centering
       \scriptsize
    \caption{Geometric transformations, called \texttt{TransGeo}.}
    \begin{tabular}{l|p{0.55\textwidth}|l}
    \hline
        Transformation & Description  &  Parameter Setting\\\hline
        RandTranslate X& Translate the image horizontally by $\lambda\times$image width.
         & $\lambda \in U(-0.1,0.1)$\\\hline
         RandTranslate Y& Translate the image vertically by $\lambda\times$image height.
         & $\lambda \in U(-0.1,0.1)$\\\hline
          RandRotate Y& Rotates the image by $\theta$ degrees.
         & $\theta \in U(-30^\circ,30^\circ)$\\\hline
         RanShear X & Shears the image along the horizontal axis with rate $R$. & $R\in U(-0.480,0.480)$ \\\hline
         RanShear Y & Shears the image along the vertically axis with rate $R$. & $R\in U(-0.480,0.480)$ \\\hline
    \end{tabular}
    \label{tab:TransGeo}
\end{table}

\clearpage
%
%
\bibliographystyle{splncs04}
\bibliography{egbib}